\documentclass[acmsmall,screen,nonacm,review=false,timestamp=false]{acmart}
\renewcommand\footnotetextcopyrightpermission[1]{} 
\usepackage[plain]{fancyref}
\usepackage[draft=true]{minted} 
\usepackage{color}
\usepackage{hyperref}           
\hypersetup{
    colorlinks=true,
    linkcolor=blue,
    filecolor=red,      
    urlcolor=magenta,
    breaklinks=true,            
}
\usepackage{breakurl}           

\usepackage{lipsum}
\usepackage{multirow}
\usepackage{caption}
\usepackage{subcaption}
\usepackage{tikz}
\usetikzlibrary{shapes,decorations,arrows,calc,arrows.meta,fit,positioning}
\tikzset{
    -Latex,auto,node distance =0.6 cm and 0.6 cm, thick, line width = 1.5,
    state/.style ={circle, draw, thick, minimum width = 0.8 cm, line width=1pt},
    point/.style = {circle, draw, inner sep=0.04cm,fill,node contents={}},
    bidirected/.style={Latex-Latex,dashed},
    el/.style = {inner sep=2pt, align=left, sloped},
    every picture/.style={line width=1pt}
}
\newcommand{\CI}{\mathrel{\perp\mspace{-10mu}\perp}}
\newcommand{\nCI}{\centernot{\CI}}
\begin{document}

\title{Understanding Disparities in Post Hoc Machine Learning Explanation}
\pagestyle{plain}

\author{Vishwali Mhasawade}
\affiliation{%
  \institution{New York University}
      \country{USA}
}
\email{vishwalim@nyu.edu}

\author{Salman Rahman}
\affiliation{%
  \institution{New York University}
      \country{USA}
  }
\email{salman@nyu.edu}

\author{Zoe Haskell-Craig}
\affiliation{%
  \institution{New York University}
      \country{USA}
  }
\email{zjh235@nyu.edu}

\author{Rumi Chunara}
\affiliation{%
  \institution{New York University}
  \country{USA}}
\email{rumi.chunara@nyu.edu}
\renewcommand{\shortauthors}{Mhasawade et al.}

\begin{abstract}
Previous work has highlighted that existing post-hoc explanation methods exhibit disparities in explanation fidelity (across “race” and “gender” as sensitive attributes), and while a large body of work focuses on mitigating these issues at the explanation metric level, the role of the data generating process and black box model in relation to explanation disparities remains largely unexplored.  Accordingly, through both simulations as well as experiments on a real-world dataset, we specifically assess challenges to explanation disparities that originate from properties of the data: limited sample size, covariate shift, concept shift, omitted variable bias, and challenges based on model properties: inclusion of the sensitive attribute and appropriate functional form. Through controlled simulation analyses, our study demonstrates that increased covariate shift, concept shift, and omission of covariates increase explanation disparities, with the effect pronounced higher for neural network models that are better able to capture the underlying functional form in comparison to linear models. We also observe consistent findings regarding the effect of concept shift and omitted variable bias on explanation disparities in the Adult income dataset. Overall, results indicate that disparities in model explanations can also depend on data and model properties. Based on this systematic investigation, we provide recommendations for the design of explanation methods that mitigate undesirable disparities. 

\end{abstract}



\maketitle

\section{Introduction}\label{sec:introduction}

Machine learning models are increasingly being proposed for and utilized in many societal areas such as healthcare, law, education, and policy-making \citep{tuggener2019automated,wu2023bloomberggpt, adeshola2023opportunities,zhao2023machine,dressel2018accuracy,subbaswamy2020development}. Particularly in their role as predictive tools, these models often are utilized with a `black box' nature. This characteristic can obscure the understanding of the underlying mechanisms driving their predictions \citep{rudin2019stop, dietvorst2015algorithm}. The lack of transparency raises concerns about the reliability of these models in situations where safety is a critical factor \citep{ghassemi2020review, ahmad2018interpretable, bussmann2021explainable}. For example, in the context of healthcare, a proposed application of machine learning models may be to determine patient treatment plans. However, as these algorithms may sometimes lead to biased predictions for disadvantaged groups, clear insights into the factors influencing the machine learning model decisions are needed \citep{chen2021ethical}. 

To address the lack of transparency in machine learning models, the field has seen a development towards Explainable AI (XAI), which focuses on creating methods that can explain the workings of these black box models \citep{linardatos2020explainable, burkart2021survey, dovsilovic2018explainable}. Among the approaches in Explainable AI, the development of simpler models that emulate the black box models' behaviors has widespread adoption in the field \citep{burkart2021survey}. This approach, known as post hoc explanation, involves developing a local model that provides explanations for individual predictions. Such explanation models are proposed for use as standalone tools, providing global explanations that shed light on the overall behavior and patterns within the black box model \citep{tan2018distill}, or to explain individual predictions, offering insights into the decision-making process of individual instance \citep{ribeiro2016should, lundberg2017unified}. 

Post hoc explanation methods are broadly classified into four categories: counterfactual \citep{wachter2017counterfactual}, rule-based \citep{ribeiro2018anchors}, perturbation-based \citep{ribeiro2016should, lundberg2017unified, plumb2018model, slack2021reliable}, and gradient-based \citep{selvaraju2017grad, smilkov2017smoothgrad}. Counterfactual explanations are computationally expensive \citep{karimi2020model} due to the demanding nature of searching for counterfactual instances in high-dimensional feature spaces. Additionally, some counterfactual suggestions may not be feasible in real-world contexts, as the changes they propose might be impractical to achieve \citep{laugel2019dangers}. Rule-based methods, on the other hand, can sometimes generate complex and hard-to-understand rules, particularly with high-dimensional data \citep{ribeiro2018anchors}.  Further, finding the most effective rule can also be computationally intensive, especially for complex models. Gradient-based methods have their limitations too; they are sensitive to noise in the input space \citep{yosinski2015understanding}, ineffective in detecting spurious correlations \citep{adebayo2022post}, are commonly applied to unstructured data like images, and sometimes produce visually similar explanations for different classes \citep{adebayo2018sanity}. Though each type of method has limitations, our focus is on perturbation-based post hoc explanation methods, especially LIME (Local Interpretable Model-agnostic Explanations), due to its widespread adoption for tabular data \citep{allgaier2023does} and previous use to highlight disparities in post hoc explanation methods \citep{dai2022fairness,balagopalan2022road}. 

Specifically, recent studies have revealed disparities in the fidelity of post hoc explanation methods, i.e., how accurately the post hoc explanation methods replicate the nature of the black box model, particularly when analyzed across data from different `gender' and `race' groups \cite{dai2022fairness, balagopalan2022road}. To address this, \citet{dai2021will} proposed a fairness-preserving approach for LIME, which includes a fairness constraint in the LIME objective function. This approach builds upon previous studies that enhanced fairness in machine learning methods through similar constraints \citep{zafar2017fairness, kamishima2012fairness} but did not extensively discuss how the fairness constraint can improve the fidelity of LIME explanations. Concurrently, \citet{balagopalan2022road} developed a robust LIME explanation model using the `Just train twice' methodology \citep{liu2021just}. However, the fidelity improvement with this enhancement was demonstrated only in certain cases; Adult, \MakeUppercase{lsac}, and \MakeUppercase{MIMIC} datasets and only for neural network methods \citep{balagopalan2022road}. 

A key point, however, is that fairness issues in machine learning models (i.e., prediction models) are a multifaceted problem that can manifest at the level of the data, the black box models, or their interpretation (e.g., via explanation methods) \citep{gebru2021datasheets, barocas-hardt-narayanan}. While efforts to-date in explanation methods have predominantly focused on improving the post hoc methods \citep{dai2021will, balagopalan2022road}, this overlooks the role of data and black box models in generating unfair explanations. Given the evidence that sample size, covariate shift, concept shift, and omitted variables can affect model prediction accuracy and lead to disparities in black box model performance, we investigate how these characteristics of the data and the model development process affect explanation disparities. 
Indeed, sample size imbalance has been linked to bias in prediction \citep{kleinberg2022racial} and calibration models \citep{ricci2023towards}.
Moreover, limited samples of a certain subgroup are known to affect model performance and the ability to generalize for the specific subgroup \citep{namkoong2023diagnosing, chen2023algorithmic}.
A related source of algorithmic unfairness is disparately missing data across subgroups \citep{martinez2019fairness, wang2021analyzing}, which can result in both an imbalance and a sample that is not representative of the true distribution of the target population, resulting in a distribution shift for some subgroups \citep{pessach2023algorithmic}.
Specifically, covariate shift is known to affect black box model performance across subgroups disparately \citep{singh2021fairness}.
Concept shift in the outcome, that is, where the conditional distribution of the outcome given the covariates varies across subgroups, also may have an effect on the quality of black box prediction and can lead to disparities \citep{namkoong2023diagnosing}. Lastly, omitting variables that have a direct effect on the outcome that is not completely mediated by other covariates may also lead to disparities in black box predictions across subgroups \citep{mhasawade2021causal}. However, there is little evidence of how characteristics of limited sample size, covariate shift, concept shift, and omitted variable bias will influence the quality of the explanation methods with respect to the test distribution.
Considering that the four above-mentioned characteristics have the potential to introduce disparities in black box model predictions, it is pertinent to investigate if and how these disparities in black box model predictions can lead to explanation disparities.  
This is motivated by the inherent nature of the explanation methods by which they are expected to mimic the nature of the black box model. 
Consider the LIME explanation method where the explanation produced by LIME at a local point \( x \) is obtained by the following generic formula: \( \xi(x) = \arg \min_{g \in G} L(f, g, \pi_x) + \Omega(g) \), where \( f \) is our real function (the black box model), \( g \) is a linear surrogate function we use to approximate \( f \) in the proximity of \( x \) and \( \pi_x \) defines the locality. Data and modeling characteristics that affect black box model performance may also have similar effects on the explanation quality as well. In sum, we explore the following data issues known to affect black box model performance but relatively unexplored in the case of model explanation disparities, 1) limited samples, 2) covariate shift, 3) concept shift, and 4) omitted variable bias.

The remainder of this paper is structured as follows: in Section \ref{sec:related_work}, we discuss related work on post hoc explanation methods and the fidelity of these approaches. In Section \ref{sec:objective-dgp}, we outline the objectives and motivate the data-generating process for each objective of our research. In Section \ref{sec:methods}, we describe the explanation quality metrics and experimental setup for the synthetic and real-world data experiments. In Section \ref{sec:results}, we present the results, and finally we discuss implications of these findings as well as synthesized recommendations based on them in Section \ref{sec:discussion}. 

\section{Related Work}\label{sec:related_work}

\noindent \textbf{Challenges in the Use of the Popular Post Hoc Explanation Method LIME.} In the realm of Explainable AI (XAI), a distinction exists between models designed for inherent interpretability \citep{bien2009classification, lakkaraju2016interpretable, caruana2015intelligible}, such as decision trees \citep{letham2015interpretable} and rule lists \citep{zeng2017interpretable, wang2015falling}, and those employing post hoc explanation methods \citep{ribeiro2016should}. Given their higher accuracy, complex models like deep neural networks are frequently preferred in real-world settings, necessitating the use of post hoc methods to elucidate their prediction. Among post hoc explanation techniques, the Local Interpretable Model-agnostic Explanations (LIME) method stands out as a widely used method, particularly for explaining black box models applied to tabular data \citep{allgaier2023does}. LIME is also valued for its model-agnostic nature, its capacity to provide local explanations, and its relative simplicity \cite{ribeiro2016should}.

However, several challenges with the use of LIME have been noted. This method relies on perturbations, which introduce computational demands, particularly in models with numerous features \citep{ribeiro2016should}. Additionally, the fidelity of LIME's explanations is sensitive to adjustments in hyperparameters, including the number of perturbed samples, kernel width, and regularization parameters. Recent research efforts have acknowledged LIME's limitations and proposed improvements. For example, S-LIME introduces frameworks for generating more stable and consistent explanations across various perturbations \citep{zhou2021s}. Despite several issues associated with LIME, systematic studies exploring performance degradation, especially concerning disadvantaged groups, are lacking. Our research aims to be the pioneering effort in investigating why post hoc explanation methods like LIME exhibit disparate explanations across different subgroups. \\

\noindent \textbf{Disparities In Post Hoc Explanation Methods and Efforts to Mitigate.} Recent research has delved into the exploration of race and gender-based disparities in a range of post hoc explanation methods, including LIME, SHAP, SmoothGrad, IntGrad, VanillaGrad, and Maple \citep{dai2022fairness, dai2021will}. These studies have utilized several datasets, such as German Credit, Student Performance, Adult, and COMPAS. Further, \citet{balagopalan2022road} conducted a study revealing explanation disparities among race and gender in both local explanation methods (LIME and SHAP) and global methods (Generalized Additive Model (GAM) and sparse decision tree (Tree)). Their research used Adult, LSAC, MIMIC, and Recidivism datasets covering four critical domains: finance, college admissions, healthcare, and the justice system. In terms of efforts to improve the fairness of explanation methods, \citet{balagopalan2022road} demonstrated that balanced samples between the subgroups and robust training for local and global explanation methods can improve the fidelity gap, which refers to how well an explanation model approximates the behavior of a black box model \citep{balagopalan2022road}. For local explanation methods, the authors trained a fairer explanation model using the Just Train Twice (JTT) methodology \citep{liu2021just}. Although improvements were noted for neural network models applied to Adult, LSAC, and MIMIC datasets, the authors did not see improvements in explanation fairness for the Recidivism dataset. As the properties of these datasets were not investigated, it is not yet clear why improvements were seen in some datasets but not others or under what conditions these methods may improve explainability. \citet{balagopalan2022road} also observed that fidelity gaps depend on the representation of the data; they train black box models with features that have no mutual information with respect to the sensitive attribute and observe that fidelity gaps decrease. Although this provides insight into one specific property of data, how much information about the sensitive attribute is available in the data representation, the authors suggest further investigation about other data properties, which forms the focus of this work. In parallel, \citet{dai2022fairness} proposed a method to generate fairness-preserving explanations by adding a penalty term to the LIME objective function \citep{dai2021will}, an approach similar to creating fair machine learning algorithms \citep{zafar2017fairness}.

While existing studies effectively highlight disparities and propose fair post hoc explanation methods, they predominantly concentrate on the outputs of the explanation methods. The role of data and black-box models in these disparities are not carefully examined though both the data used and the nature of the black-box models can be significant sources of disparity \citep{barocas-hardt-narayanan}.

\section{Data Generating Process and Objectives} 
\label{sec:objective-dgp}

\subsection{Data Generating Process}
Here, we describe the data-generating process (DGP) for assessing the reasons for disparities in model explanations in line with previous work using simple causal graphs for systematic fairness assessments \citep{singh2021fairness,mhasawade2021causal,subbaswamy2018counterfactual}. 
We refer to the outcome as $Y$, a binary variable that takes a value of $0$ or $1$. We consider a sensitive attribute $A$, such as race or gender, for which we represent the disadvantaged group as $A = 0$ and the advantaged group as $A = 1$. $A$ is associated with the independent covariate $L$. 
Two attributes have a direct effect on the outcome $Y$, $C$, and $L$, where $L$ mediates a part of the effect of $C$ on $Y$. $C$ has a direct effect on $Y$ ($C \rightarrow Y$) and an indirect effect through $L$ ($C \rightarrow L \rightarrow Y$).
The relationship between these variables is represented by the causal directed acyclic graph (DAG) in Figure \ref{fig:dags}(a).  
In our causal graph, the covariates and the sensitive attribute affect the outcome either through other covariates (i.e., $(A \rightarrow L \rightarrow Y)$) or directly (i.e., $C \rightarrow Y$). 
 
We consider two setups; in the first, presented in Figure \ref{fig:dags}(a), $A$ has an effect on $Y$ only through $L$,  and in the second, in Figure \ref{fig:dags}(b) $A$ affects the relationship between $L$ and $Y$. The second setting allows us to assess non-linear complex functional forms between $A$ and $Y$.
We use these DGPs to represent the underlying relationship between the variables in the general population, from which we will draw samples to form training and testing datasets. 

\subsection{Objective 1: effect of sample size of disadvantaged group data used for training}

In order to investigate the effect of sample size imbalance on disparities, we consider a scenario where we vary the proportion of the sample size of the disadvantaged group ($A =0 $) from 5\% to 50\% of the training sample and accordingly vary the proportion of the advantaged group ($A=1$) from 95\% to 50\% of the training sample.
To isolate this effect from non-random sampling of data \citep{kallus2018residual}, we assume that there is no distribution shift in the predictors $L$ and $C$ between the training and test distributions.
It is important to note that the probability of the outcome $Y$, given the predictor $L$, $P(Y = y | L, C)$, remains independent of the sample size of $A=0$ since $Y \CI A \mid L, C$ and we randomly vary the proportion of both advantaged and disadvantaged group to ensure that the training sample is a perfectly random sample of the population distribution.

The DGP for this objective is represented in figure \ref{fig:dags}(a). 
We assume that, in the general population, attribute $A$ is generated by a binomial probability with $A \in \{0, 1\} \sim \text{Binomial}(1, 0.5)$, $C$ follows a normal distribution $C \sim \mathcal{N}(0, 1)$, and $P(L)$ is dependent on both $A$ and $C$ such that $L \sim \mathcal{N}(0, 0.5) + 0.7 \times A + 0.3 \times C$. These parameters were chosen to allow for differences in the distribution of $L$ across $A$ such that $P(L\mid C) \neq P(L\mid C, A)$.
We assume that covariates $L$ and $C$ have a direct effect of a similar magnitude on the outcome, $Y$, which follows a binomial distribution $Y \sim \text{Binomial}(1, Y_p)$ with probability of as $Y_p : P(Y = 1| L, C) = \begin{cases}
0.1 & \text{if } i < 0 \\
0.9 & \text{if } i \geq 1
\end{cases}, i = 0.5 \times C - 1.5 \times L + 0.5$.

We also consider both the cases, when the black-box machine learning model includes information on the sensitive attribute $A$ and when it does not \citep{vyas2020hidden,mitchell2019model}, to assess the fairness properties of explanation metrics when the population Bayes-optimal model is not subgroup Bayes-optimal \citep{pfohl2023understanding}.

\subsection{Objective 2: effect of covariate shift in disadvantaged group data between training and test distributions}

Here we explore the effect of a covariate shift, where the training distribution for $A=0$ is not representative of the test distribution (or the population distribution). 
That is, $P_{\text{train}}(L|A = 0) \ne P_{\text{test}}(L|A = 0)$, however the conditional probabilities are consistent; $P_{\text{train}}(Y | L, A = 0) = P_{\text{test}}(Y | L, A = 0) $. 
We generate a covariate shift in $L$ for $A = 0$ by sampling for the training distribution depending on both the sensitive attribute $A$ and the covariate $L$, such that there is missing data for group $A = 0$ for all observations with $L$ below a  threshold $t$. 
In this way, we vary the overlap in the range of $L$ between the test set and the training set from $100\%$ to $20\%$.
It should be noted that the probability distribution of $Y$ given $A = 0$ is not the same in the training and test sets; $P_{\text{train}}(Y|A = 0) \ne P_{\text{test}}(Y|A = 0)$. 
As the overlap between the training and testing sets is reduced, the model has less information in the training set to learn about the association between the variables in the general population for $A=0$.
As such, we hypothesize that less overlap in the training and test distributions for the disadvantaged group will lead to greater disparities. 

The DGP for this objective is also represented in Figure \ref{fig:dags}(a). 
Similar to objective 1, the data-generating process for each variable is as follows: $ A \in \{0, 1\} \sim \text{Binomial}(1, 0.5)$, $C \sim \mathcal{N}(0, 1)$, $ L = \mathcal{N}(0, 0.5) + 0.7 \times M + 0.3 \times C$, $Y_p = P(Y = 1) = \begin{cases}
0.1 & \text{if } i < 0 \\
0.9 & \text{if } i \geq 1
\end{cases}, i = 0.5 \times C - 1.5 \times L + 0.5$, $ Y \sim \text{Binomial}(1, Y_p)$. 
Note that for both objectives 1 and 2, we evaluate model performance on a test set representative of the population. That is, the proportion of $A = 0$ in the test set is $50\%$, and the full distribution of the values for $L$ is represented.
As such, $P_{\text{train}}(A) \ne P_{\text{test}}(A)$.
We again consider both cases when $A$ is and is not included in the covariates to which the black-box model has access.

\subsection{Objective 3: effect of concept shift}

Here we examine concept shift, where $P(Y|L, A = 0) \ne P(Y|L, A = 1)$ \citep{moreno2012unifying}. 
That is the relationship between $Y$ and $L$ changes depending on $A$.
Here, we vary the magnitude of concept shift by increasing the effect of the sensitive attribute on the distribution of the outcome, that is, by varying the degree of difference between the distribution of the outcome across groups that results in $P(Y) \ne P(Y|A)$. 

The DGP for this objective is represented in figure \ref{fig:dags}(b). 
We generate the distribution of variables $A$ and $C$ in the general population following the same procedure outlined in objectives 1 and 2.
In order to capture the impact of $A$ on $P(Y|L)$, we augment the direct effect of $A$ and $L$ such that $ L \sim \mathcal{N}(0, 0.1) + 0.7 \times A + 0.3 \times C$. 
The concept shift is produced by specifying that the probability $Y_p$ depends on $i = 0.5 \times C + - 1 \times L + 1.5 \times A * L + \beta \times (1-A) * L  - 0.2$, through the step function $Y_p = \begin{cases}
0.1 & \text{if } i < 0 \\
0.9 & \text{if } i \geq 0
\end{cases}$. 
Note that the relationship between $L$ and $Y_p$ is determined by $A$ and $\beta$, where $\beta$ is the strength of the concept shift.
We consider $\beta = 1.5$ as `low' concept shift, $\beta = 0.5$ a `moderate' concept shift, and $\beta = -0.5$ a `high' concept shift.
The coefficients on $C$, $L$, and the intercept term were chosen in order to ensure an equal distribution of $Y = 1$ and $Y = 0$ in the training sample to ensure class balance.
As before, we consider the impact of including or not including information on $A$ in the black box model.

\subsection{Objective 4: effect of the magnitude of direct effect of the omitted covariate}

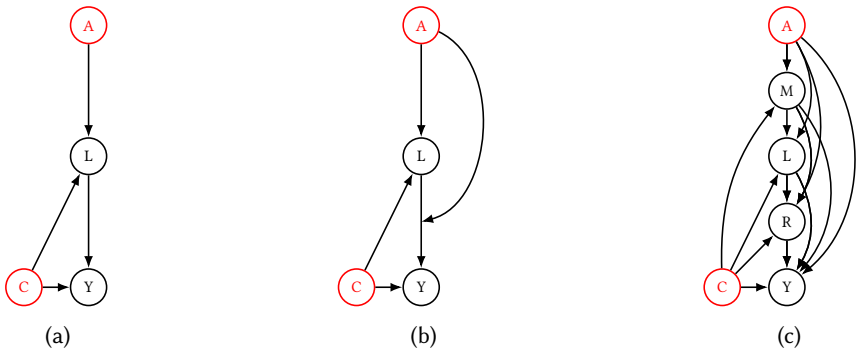
\begin{figure}[!htbp]
    \centering
    \begin{subfigure}[b]{0.3\textwidth}
        \centering
        \scalebox{0.6}{\begin{tikzpicture}
    \node[state,red] (c) at (0,0) {C};
    \node[state] (y) [right =of c] {Y};
    \node[state,opacity=0] (r) [above =of y] {R};
    \node[state] (l) [above =of r] {L};
    \node[state,opacity=0] (m) [above =of l] {M};
    \node[state,red] (a) [above =of m] {A};
    \path (a) edge (l);
    \path (c) edge (l);
    \path (c) edge (y);
    \path (l) edge (y);
\end{tikzpicture}}
        \caption{}
    \end{subfigure}
    \hfill
    \begin{subfigure}[b]{0.3\textwidth}
        \centering
        \scalebox{0.6}{\begin{tikzpicture}
    \node[state,red] (c) at (0,0) {C};
    \node[state] (y) [right =of c] {Y};
    \node[state,opacity=0] (r) [above =of y] {R};
    \node[state] (l) [above =of r] {L};
    \node[state,opacity=0] (m) [above =of l] {M};
    \node[state,red] (a) [above =of m] {A};
    \path (a) edge (l);
    \path (c) edge (l);
    \path (c) edge (y);
    \path (l) edge (y);
    \path (l) -- coordinate[midway] (x) (y);
    \path (a) edge[bend left=70] (x);

\end{tikzpicture}}
        \caption{}
    \end{subfigure}
    \hfill
    \begin{subfigure}[b]{0.3\textwidth}
        \centering
        \scalebox{0.6}{\begin{tikzpicture}
    \node[state,red] (c) at (0,0) {C};
    \node[state] (y) [right =of c] {Y};
    \node[state] (r) [above =of y] {R};
    \node[state] (l) [above =of r] {L};
    \node[state] (m) [above =of l] {M};
    \node[state,red] (a) [above =of m] {A};
    \path (a) edge (m);
    \path  (m) edge (l);
    \path (l) edge (r);
    \path (l) edge (r);
    \path (r) edge (y);
    \path  (r) edge (y);
    \path (c) edge [bend left=20] (m);
    \path (c)  edge (l);
    \path (c) edge (r);
    \path (c) edge (y);
    \path (a) edge (m);
    \path (m) edge [bend left=30] (r);
    \path  (m) edge [bend left=30] (r);
    \path  (m) edge [bend left = 40] (y);
    \path (l)  edge [bend left = 30] (y);
    \path (l) edge [bend left = 30] (y);

    \path (a) edge [bend left = 30] (l);
    \path (a) edge [bend left = 30] (r);
    \path (a) edge [bend left = 50] (y);


\end{tikzpicture}}
        \caption{}
    \end{subfigure}
    \caption{Causal DAG for the synthetic datasets (a, b) and the Adult dataset (c). In (a) is the causal graph describing the data-generating process (DGP) for objectives 1, 2, and 4, (b) is the causal graph for the DGP for objective 3 (concept shift). The concept shift is represented as the arrow showing the effect of $A$ on the relationship between $L$ and $Y$, such that $P(Y|L, A = 0)  \ne P(Y|L, A = 1)$. In (c) we consider gender as the sensitive attribute of interest. $A$ and $M$ represent gender and marital status, respectively. $C$ is age and nationality, $L$ is the level of education, $R$ corresponds to the working class, occupation, and hours per week, and $Y$ is the income class.}
    \label{fig:dags}
\end{figure}

Finally, we test the impact of omitting a covariate, $C$, that has a direct effect on the outcome in the black box model on explanation disparities.
Following the data-generating process represented in Figure \ref{fig:dags}(a), the distribution of $A$ and $C$ are generated as before, $L \sim \mathcal{N}(0, 0.5) + 0.3 \times A + 0.3 \times C$, and as before, $ Y \sim \text{Binomial}(1, Y_p)$.
We vary the direct effect of the attribute, $C$, on the outcome that is not mediated by other covariates, such that $Y_p \sim \alpha \times C + L - 0.2$ where $\alpha \in \{ 0, 0.5, 1, 1.5\}$. 
We assess the disparities resulting from not including the variables $C$ in model training (or test).



\section{Methods}\label{sec:methods}
\subsection{Notation}
From the underlying population we draw a dataset $\mathcal{D} = \{(x_1,y_1), (x_2, y_2), \cdots, (x_n,y_n)\}$ comprising $n$ observations. 
Each observation consists of a $d$-dimensional feature vector $x_i \in \mathbb{R}^d$ for the $i^{th}$ data point in $\mathcal{D}$, along with a corresponding binary class label $y_i \in \{0,1\}$. 
We consider a machine learning model $f:\mathbb{R} \rightarrow \{0,1\}$, such as logistic regression (LR) or a neural network (NN), trained on samples from $\mathcal{D}$. 
For a given instance $x_i$ and machine learning model $f$, a local explanation method is denoted as $E:(x_i, f) \rightarrow \psi \in \mathbb{R}^d$, where $\psi$ represents the output vector of feature importance. 
The local model, $E$, is designed to mimic the behavior of $f$ in the vicinity of $x_i$ \cite{zhou2021evaluating,dai2022fairness}.

\subsection{Explanation Quality Metrics}
We utilize two metrics to measure the fidelity gap across groups, as introduced by \citet{balagopalan2022road}. 
Fidelity is the degree to which an explanation model precisely reflects the predictions of a black box model. 
For a black box model $f$ and explanation model $E$, the fidelity quantifies how closely $E$ approximates the behavior of $f$. 
Mathematically, the explanation fidelity for data points $(x_i, y_i)_{i=1}^N$ is calculated as: $\frac{1}{N} \sum_{i=1}^N Q(f(x_i), E(x_i))$, where $Q$ is a performance metric such as accuracy.
This measure allows for the evaluation of fairness by illustrating the alignment between the machine-learning model and the explanation model.

\paragraph{Maximum Fidelity Gap from Average}

The Maximum Fidelity Gap from the Average measures the largest deviation in fidelity for any group from the average fidelity across all groups. 
This metric assesses the extent to which the fidelity of an explanation model for disadvantaged groups deviates from the overall average fidelity \citep{balagopalan2022road, liu2021just, craven1995extracting}. 
The maximum fidelity gap from average, $\Delta_{Q}$ is represented as follows:

 \begin{equation*}
\Delta_{Q} = \max_{j} \left[ \frac{1}{N} \sum_{i=1}^N Q(f(x_i), E(x_i)) - \frac{1}{N_j} \sum_{i:\delta_{j}^{i} =1} Q(f(x_i), E(x_i)) \right]
\end{equation*}

where $Q$ represents performance metric such as Accuracy, $N$ is the total number of data points, $\delta_{j}^{i} = 1$ indicates that point $x_i$ belongs to the $j$-th group, and $N_j$ is the number of data points where $\delta_{j} = 1$. 
We specifically focus on the maximum fidelity gap from the average for the `Accuracy' performance metric following the performance metric used by \citet{balagopalan2022road}, where we assess the accuracy between the predictions of the black box model $f(:)$ and the explanation method $E(:)$ and denote it as $\Delta_{Acc}$.

\paragraph{Mean Fidelity Gap Amongst Subgroups} 

This metric illustrates the average difference in fidelity between groups.
Within the Mean Fidelity Gap, performance metrics such as AUROC, Residual Error, and Accuracy may be used to quantify disparities between black box model predictions and their explanations. 
This metric is computed as follows \citep{balagopalan2022road}:

\begin{equation*}
\Delta_{Q}^{group} = \frac{2}{G(G-1)} \sum_{p=1}^{G} \sum_{j=p+1}^{G} |Q_p - Q_j|
\end{equation*}
with
\begin{equation*}
Q_p = \frac{1}{N_p} \sum_{i:\delta_{p}^{i} =1} Q(f(x_i), E(x_i))
\end{equation*}

where $G$ is the total number of groups, $Q_p$ and $Q_j$ are the performance metrics for the $p^{th}$ and $j^{th}$ groups respectively, $\delta_p$ denotes the $p$-th group, and $N_j$ is the number of data-points in $\delta_p$. 
As described above, the $Q$ metric includes 
Accuracy, denoted by 
$\Delta_{Acc}^{group}$.
We use Accuracy as the performance metric for both explanation quality metrics to compare how accurate the explanations for the disadvantaged group are in comparison to overall and how accurate the explanations for the disadvantaged group are in comparison to the advantaged group.

\subsection{Experimental Setup}

\noindent \textbf{Simulation and Real-world Data.} The simulated data, as outlined in Section \ref{sec:objective-dgp}, is constructed to reflect a specific causal structure. 
This dataset comprises 20,000 data points featuring a sensitive attribute $A$ (such as gender), covariates $L$ and $C$, and a binary outcome $Y$. 
Specifically, the covariate $L$ mediates a part of the effect of $C$ on $Y$.
The process for generating this data is detailed in Section \ref{sec:objective-dgp}.
We assess the quality of the explanations generated with respect to the sensitive attribute $A$. 

In addition to the simulated data, we employ the widely used Adult dataset for real-world analysis \cite{dua2017uci,lichman2013uci}. 
We chose this dataset due to the importance of the dataset in developing and evaluating post-hoc explainability methods  \citep{sangroya2020guided,velmurugan2021developing, hase2020evaluating}, the general popularity of the dataset for fairness analysis \citep{fabris2022algorithmic,begley2020explainability, balagopalan2022road,dai2022fairness} and the availability of a well studied causal DAG \cite{mhasawade2021causal,chiappa2019path}.
This previously used causal graph is shown in Figure \ref{fig:dags}(c). 
For the task using the Adult dataset, the goal is to predict whether an individual’s income is above or below \$50,000. 
The data consist of age, working class, education level, marital status, occupation, relationship, race, gender, capital gain and loss, working hours, and nationality variables for 48842 individuals.
In Adult, disparities in explanation quality (maximum fidelity gap and mean fidelity gap between groups using accuracy and error) have been found for both logistic regression and neural network black box models with respect to gender \citep{balagopalan2022road}, but the source of these disparities has not been investigated.
Following \cite{mhasawade2021causal}, we consider the variable `hours-per-week` as one of the predictors with a direct effect on the outcome of interest, income $Y$ represented by $L$ in the synthetic experiment.
All other covariates that may be associated with both `hours-per-week` and income are represented as $C$ in the synthetic experiment, in particular age and nationality. 
We consider gender as the sensitive attribute of interest (‘$A$’ in the synthetic experiment) as has been done in previous studies on algorithmic fairness\citep{mhasawade2021causal,balagopalan2022road}, and considering `males' as the disadvantaged group since we observe that it is easier to predict the outcome for the advantaged group `females' than it is to predict the outcome for the disadvantaged group `males.'

Moreover, for this dataset and task, we detect a statistically significant concept shift ($p \leq 0.1$) of gender ($A$) on the relationship between `hours-per-week' ($L$) and the outcome ($Y$).
The concept shift was tested using logistic regression with an interaction term, such that (logit(income) = $\beta_1$sex + $\beta_2$``hours-per-week" + $\beta_3$sex ``hours-per-week").
Furthermore, following the proposed causal graph for the Adult dataset, we use `nationality' as the omitted variable ($C$ in the synthetic experiment), which has a direct effect on $Y$ along with indirect effects on $Y$ through other covariates. We examine these effects further by performing the following experiments, which match the experiments we performed with synthetic data. 
We analyze the effect of the sample size of the disadvantaged group (objective 1) by further restricting the proportion of females in the training dataset from  10\% to 100\% (with a 10\% increment). 
We also perform experiments examining the effect of non-overlap between training and test distribution of `hours-per-week` (objective 2) by limiting the observations of males in the training dataset to individuals working less than 100, 80, 60, 40, and 20 hours per week and thus introducing covariate shift in `hours-per-week' for males.
We also check the impact of concept shift alone by ensuring the training set is 50\% females and 50\% male (objective 3). This ensured that we had an equal representation of both groups in the training set.
We examine the effect of omitting a covariate that has a direct effect on the outcome (objective 4) by omitting a) gender, b) nationality, and c) both from the black box model. 

\noindent \textbf{Machine Learning Model.}  Our study examines both logistic regression (LR) and neural network (NN) models as the underlying functions in the black box model which are implemented using the PyTorch framework. 
This selection enables us to compare the impacts of a simpler, linear model and a more complex, flexible model on the quality of explanations, aligning with methodologies used in previous studies \cite{dai2022fairness, balagopalan2022road}.
The neural network architecture consists of four layers linked by ReLU activation functions and concludes with a final layer with a sigmoid activation function for output, mirroring the setup used for assessing disparities by \citet{dai2022fairness}.
The layer configurations are as follows: the first linear layer maps input features to 50 outputs, succeeded by ReLU activation.
The subsequent layers expand the dimensionality (50 to 100, and then 100 to 200), each followed by ReLU activations. 
The final linear layer condenses these 200 inputs to a single output, processed through a sigmoid activation function.
We utilize the Adam optimizer with a weight decay of $1e^{-4}$ and train the model using Binary Cross-Entropy Loss over 100 epochs.
We represent the NN models that do not include the sensitive attribute in model training as NN$_{\not \mathbf{A}}$ and the NN models that include the sensitive attribute as NN$_{A}$ and similar for the LR models as LR$_{\not \mathbf{A}}$, and LR$_{A}$, respectively.

\noindent \textbf{Explanation Method.}  As motivated above, our focus is on LIME for the explanation model, chosen for its extensive application and documented disparities in previous studies \cite{allgaier2023does, dai2022fairness, balagopalan2022road}. 
LIME operates by constructing a local surrogate model to interpret specific data points, thereby shedding light on the prediction of the underlying complex model.
LIME generates a dataset of perturbations by altering the features of a specific instance, creating a range of variations. 
The original machine learning model is then used to obtain predictions for the dataset comprising of the perturbed instances.
The perturbed instances are weighted according to their similarity to the original instance by comparing the distance to the original instance from which the perturbations are obtained. 
Subsequently, a simpler, interpretable linear model is trained on this weighted, perturbed dataset \citep{ribeiro2016should}. 
The objective of this simpler model is to approximate the complex model's predictions in the vicinity of the selected instance. 
The explanation is obtained from the features of this simpler model, identifying the key covariates influencing the specific prediction. 
The implementation of LIME utilizes LimeTabularExplainer from the `lime' package in Python, which operates on the training data without discretizing continuous features. 
The result from this implementation is 1000 perturbed samples for informing the explanation of each instance, considering all dataset features for generating explanations for each test instance.

\noindent \textbf{Settings and Implementation Procedures.}
The datasets $\mathcal{D}$ are divided into training $\mathcal{D}_{\text{train}}$ and testing 
 $\mathcal{D}_{\text{test}}$ sets, comprising $70\%$ and $30\%$ of the data, respectively. 
 The black box model $f$ is trained using $\mathcal{D}_{\text{train}}$. 
 The test dataset $\mathcal{D}_{\text{test}}$ is further segmented into two groups $\mathcal{D}_{\text{test}}^{A=1}$ for group 1 and  $\mathcal{D}_{\text{test}}^{A=0}$ for group 0. 
 We generate explanations using LIME for both groups.
 The fidelity of these explanations is assessed by comparing the predictions from the black box model $f$ and the explanation model $E$, using the fidelity metrics defined in Section 4.
 To evaluate the consistency of predictions between the black box model $f$ and the explanation model $E$, we conduct five trials, each with different random seeds.

\section{Results}\label{sec:results}

\subsection{Explanation Disparities in Synthetic Simulations}
\begin{figure}[h]
    \centering
    \begin{subfigure}[b]{0.45\textwidth}
    \centering
    \includegraphics[scale=0.17]{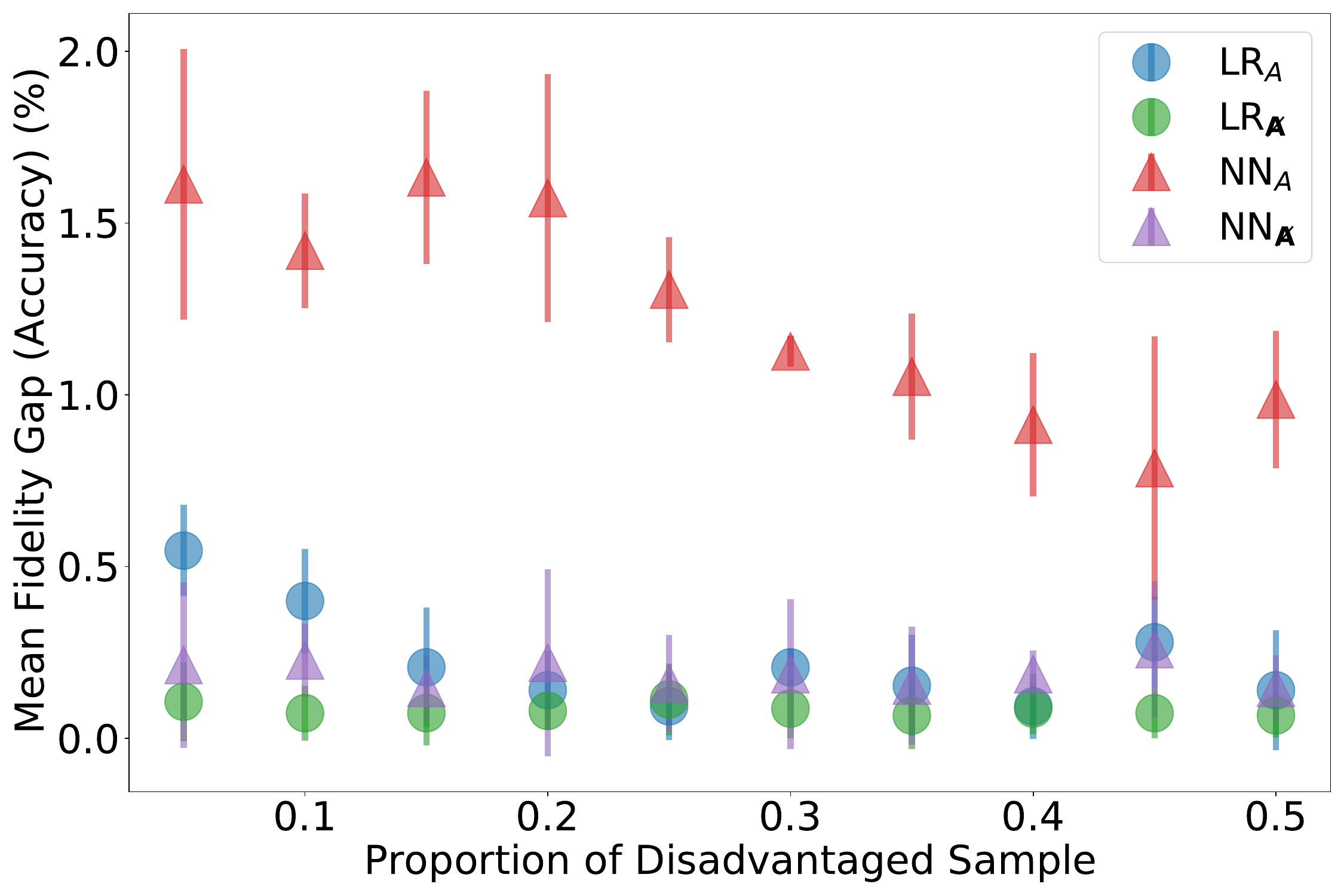}
    \caption{}
    \end{subfigure}
    \hfill
    \begin{subfigure}[b]{0.45\textwidth}
    \centering
\includegraphics[scale=0.17]{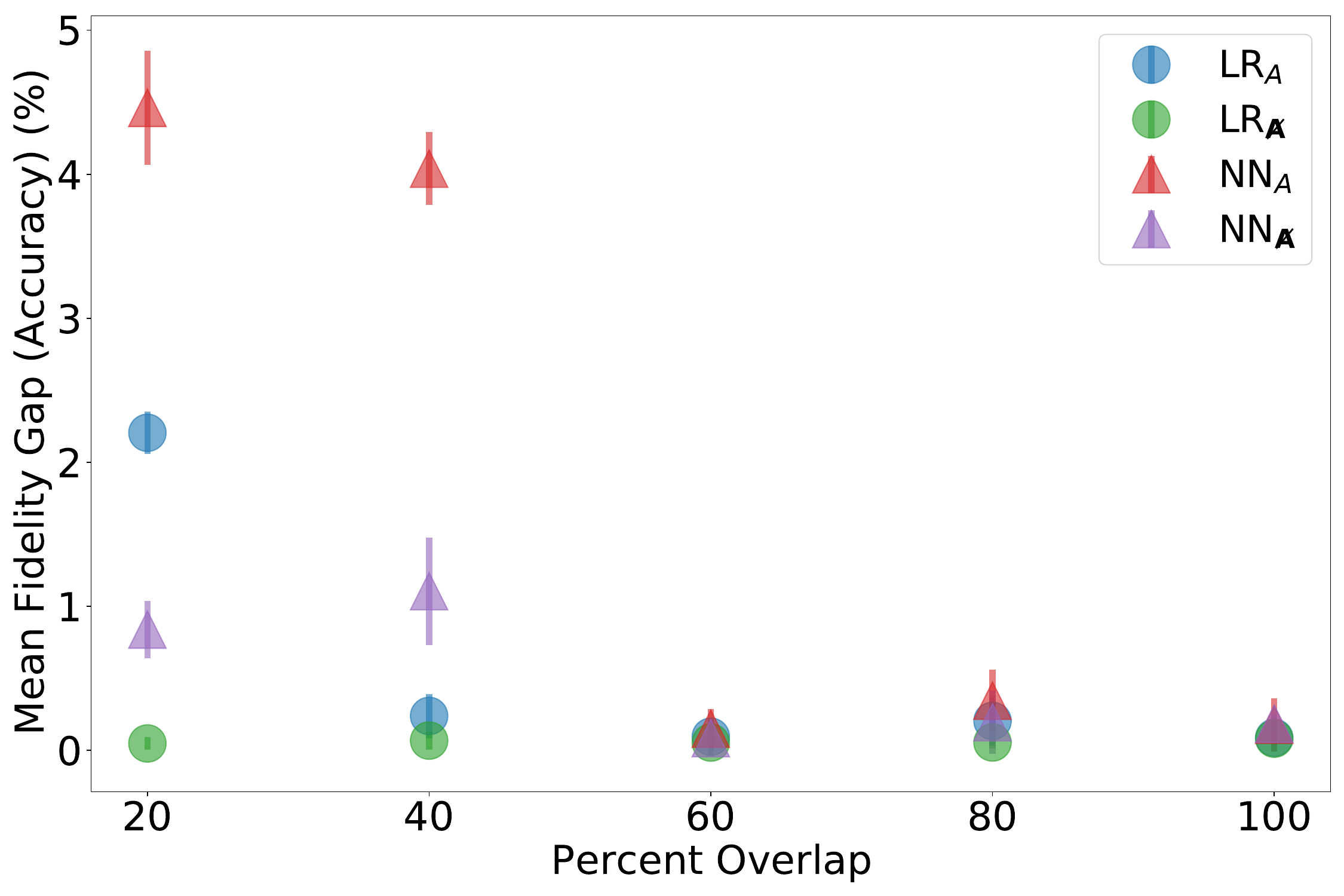}
    \caption{}
    \end{subfigure}
    \hfill
    \begin{subfigure}[b]{0.45\textwidth}
    \centering
    \includegraphics[scale=0.17]{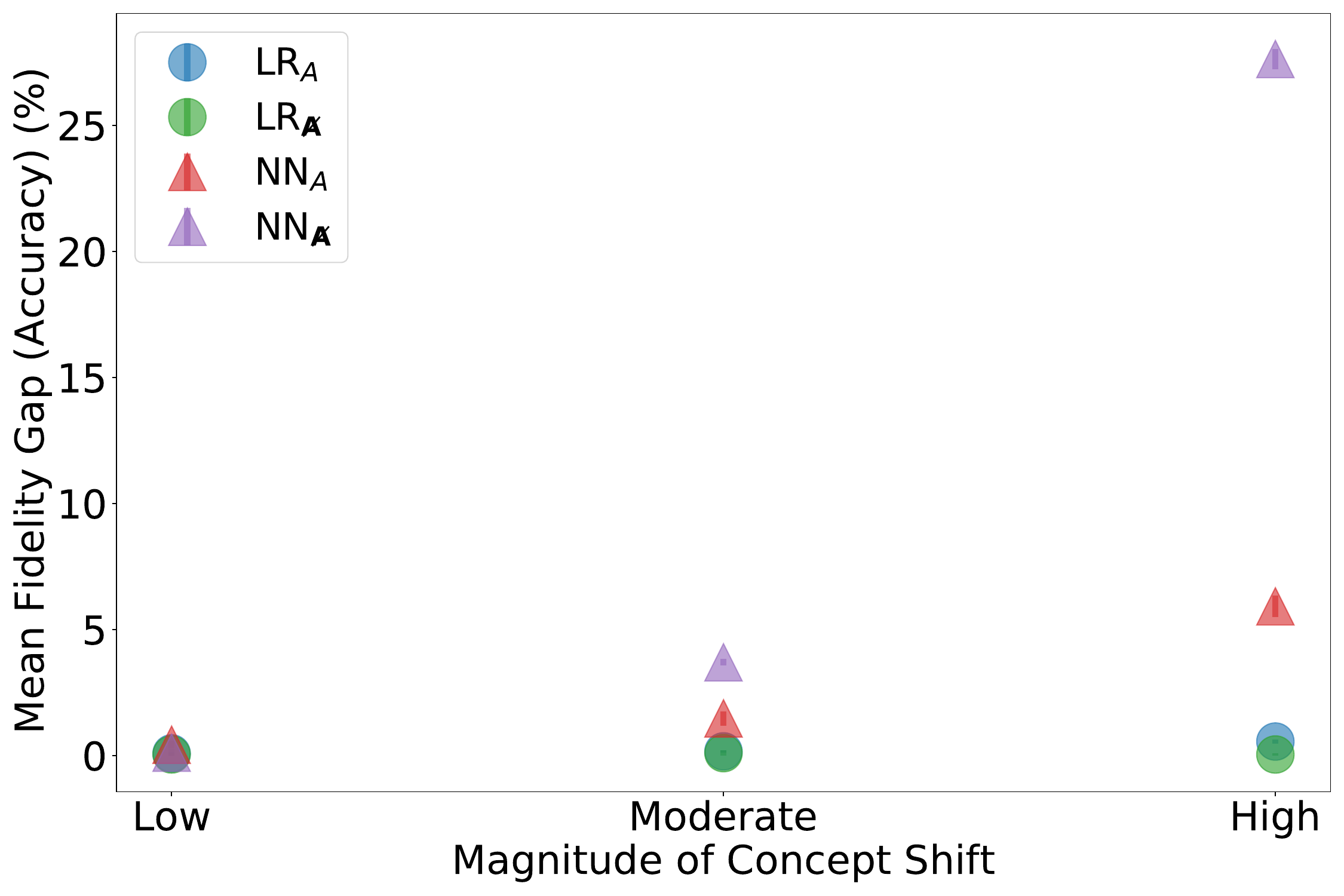}
    \caption{}
    \end{subfigure}
    \hfill
    \begin{subfigure}[b]{0.45\textwidth}
    \centering
    \includegraphics[scale=0.17]{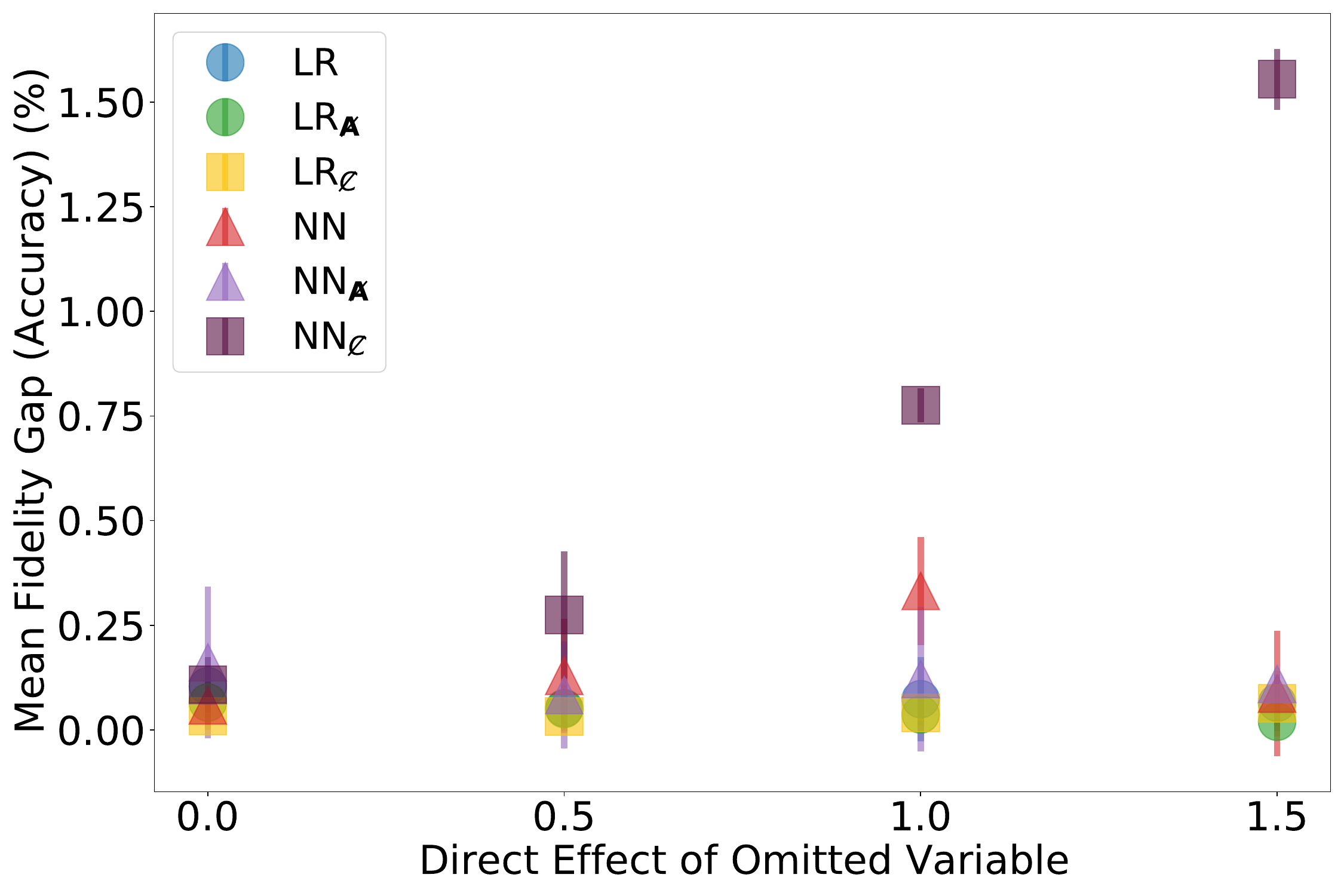}
    \caption{}
    \end{subfigure}
    \caption{Percent Mean Fidelity Gap (Accuracy) of LIME applied to models built on the synthetic datasets generated for (a) objective 1 - sample size, (b) objective 2 - covariate shift, (c) objective 3 - concept shift, and (d) objective 4 - omitted variables. In (a), we vary the proportion of the disadvantaged group in the training set sample. In (b), we introduce a covariate shift for the disadvantaged group, shifting the overlap between the train and test distributions; and in (c), we vary the magnitude of the concept shift. In (d), we adjust the strength of the direct effect of the omitted variable $C$. The models that are considered are LR with $A$, LR$_{A}$ in blue, LR without $A$, LR$_{\not \mathbf{A}}$ in green, NN with $A$, NN$_{A}$ in red, and NN without $A$, NN$_{\not \mathbf{A}}$ in violet, LR without $C$, LR$_{\not C}$ in yellow, and NN without $C$, NN$_{\not C}$ in plum.  Circles represent linear models, and triangles represent neural network models. Notice that the magnitude of the mean fidelity gap is much larger under conditions of (b) covariate shift and (c) concept shift than either (a) sample size differences or (d) omitted variables. }
    \label{fig:results_sim}
\end{figure}

\noindent \textbf{Overall findings on model complexity and inclusion of sensitive attributes.}
We observe similar characteristics across all objectives between simpler linear models (LR) and complex neural network models (NN).
In general, higher disparities in explanation metrics (Maximum Fidelity Gap, $\Delta_{Acc}$ and Mean Fidelity Gap, $\Delta_{Acc}^{group}$) are found for models that use higher complexity for the functional form (NN), in comparison to simpler models (LR).
Moreover, if the inclusion of the sensitive attribute ($A$) in training the black box model aligns with the causal structure ($Y \nCI A \mid C, L$) then $A$ needs to be included in model training, and explanation disparities are smaller than if the inclusion of group information does not align with the causal structure, (if $A$ is not included in the black box model training even though $Y \nCI A \mid C,L$). 
Specifically for objectives 1 and 2, excluding the sensitive information aligns with the causal structure, and accordingly, models that include the sensitive attribute LR$_{A}$ and NN$_{A}$ have higher disparities when compared with models that exclude the sensitive attribute, i.e., LR$_{\not \mathbf{A}}$ and NN$_{\not \mathbf{A}}$, respectively as reflected in Figures \ref{fig:results_sim}(a), \ref{fig:results_sim}(b).
Moreover, we observe that the highest magnitude of $\Delta_{Acc}^{group}$ and $\Delta_{Acc}$ is for NN and is much larger under conditions of covariate shift (4.52\%, 2.25\%) and concept shift (27.63\%, 14.12\%) than either sample size differences (1.6\%, 0.82\%) or omitted variables (1.6\%,0.89\%).
While we specifically report $\Delta_{Acc}$ and $\Delta_{Acc}^{group}$ here, we observed similar behavior across $\Delta_{AUROC}^{group}$, $\Delta_{Error}^{group}$, not reported here for brevity.

\noindent {\textbf{Objective 1: 
 Observations with respect to variation in sample size of disadvantaged group.}}\\
\noindent As the proportion of disadvantaged group samples increases, making their representation in the training set closer to the test set, explanation disparity metrics for NN$_{\not \mathbf{A}}$ and LR$_{\not \mathbf{A}}$ remain approximately consistent, illustrated in Figure \ref{fig:results_sim}(a) and Appendix  Figure \ref{fig:appendix_results_sim_maximum}(a).
For a proportion of 0.05 of the disadvantaged group in the training sample, $\Delta_{Acc}^{group}$ for NN$_{\not \mathbf{A}}$ and LR$_{\not \mathbf{A}}$ are 0.19\% and 0.23\%, respectively while for a proportion of 0.5 of the disadvantaged group in the training sample, $\Delta_{Acc}^{group}$ corresponds to 0.19\% and 0.21\%, respectively.
However, when the sensitive attribute is used for model training in the case of NN$_{A}$ and LR$_{A}$, larger model explanation disparities result.
Specifically, for 0.05 proportion of the disadvantaged group, $\Delta_{Acc}^{group}$ for NN$_{A}$ and LR$_{A}$ is 0.52\% and 1.60\%, respectively, an increase of 0.33\% and 1.37\% from NN$_{\not \mathbf{A}}$ and LR$_{\not \mathbf{A}}$.
At 0.5 proportion of the disadvantaged group, $\Delta_{Acc}^{group}$ for NN$_{A}$ and LR$_{A}$ drop to 1.1\% and 0.21\%, respectively. 
Figure \ref{fig:results_sim}(a) and Appendix Figure \ref{fig:appendix_results_sim_maximum}(a)
show these findings.
Thus, disparities decrease for models that include the sensitive attribute but remain consistent for models that exclude the sensitive attribute.
It also should be noted that the difference in black box model performance (accuracy) for the disadvantaged and advantaged groups is consistent for NN$_{\not \mathbf{A}}$ and LR$_{\not \mathbf{A}}$, but the difference decreases for NN$_{A}$ and LR$_{A}$ as the proportion of the disadvantaged group increases in the training sample, as illustrated in Appendix Figure \ref{fig:bb_sim_1}(a).

\noindent {\textbf{Objective 2: 
Variation in covariate shift for the disadvantaged group}}\\
\noindent Introducing a covariate shift in $L$, specifically for the disadvantaged group, results in the training distribution of the disadvantaged group not being representative of the test distribution of the disadvantaged group.
As the overlap between the training and test distributions of the disadvantaged group increases, overall disparities in the explanation metrics go down.
Specifically, for 20\% overlap, $\Delta_{Acc}^{group}$ for NN$_{\not \mathbf{A}}$ and LR$_{\not \mathbf{A}}$ are 0.84\% and 0.05\%, respectively, while for 100\% overlap of the disadvantaged group between the training and test distributions, $\Delta_{Acc}^{group}$ corresponds to 0.18\% and 0.08\%, respectively.
However, when the sensitive attribute is used for model training, explanation disparity in $\Delta_{Acc}^{group}$ and $\Delta_{Acc}$ is higher in comparison to when the sensitive attribute is excluded in model training, where there is incomplete overlap.
At 20\% overlap, $\Delta_{Acc}^{group}$ for NN$_{A}$ is 4.46\% and for LR$_{A}$ is 2.2\%, an increase of 3.62\% and 2.15\% from NN$_{\not \mathbf{A}}$ and LR$_{\not \mathbf{A}}$, respectively.
However, at 100\% overlap explanation disparities for NN$_{A}$ and LR$_{A}$ reduce considerably with $\Delta_{Acc}^{group}$ as 0.18\% and 0.08\%, respectively.
At 100\% overlap, explanation disparities for NN$_{\not \mathbf{A}}$ and LR$_{\not \mathbf{A}}$ are same as NN$_{A}$ and LR$_{A}$, respectively.
This is illustrated in Figure \ref{fig:results_sim}(b) for $\Delta_{Acc}^{group}$ and a similar trend is observed in other metric, $\Delta_{Acc}$
as shown in Appendix Figure \ref{fig:appendix_results_sim_maximum}(b).
The difference in black box model performance metric (accuracy) between the disadvantaged and advantaged groups is consistent for NN$_{\not \mathbf{A}}$ and LR$_{\not \mathbf{A}}$, but the difference decreases for NN$_{A}$ and LR$_{A}$ as the proportion of the disadvantaged group increases in the training sample as illustrated in Appendix Figure \ref{fig:bb_sim_1}(b).

\noindent \textbf{Objective 3: Variation in the magnitude of concept shift}\\
\noindent As the magnitude of the concept shift is increased for the disadvantaged group from low to moderate to high, disparities in model explanations for $\Delta_{Acc}^{group}$ increase as presented in Figure \ref{fig:results_sim}(c).
As the concept shift increases from moderate to high, $\Delta_{Acc}^{group}$ for LR$_{A}$ varies from 0.17\% to 0.55\% (an increase of 0.38\%)  while that for LR$_{\not \mathbf{A}}$ varies from 0.09\% to 0.04\% (a decrease of 0.05\%) .
For NN$_{A}$, $\Delta_{Acc}^{group}$ increases from 1.47\% to 5.92\% (an increase of 4.45\%). 
However, in the case of NN$_{\not \mathbf{A}}$, $\Delta_{Acc}^{group}$ increases from 3.70\% to 27.63\% (an increase of 23.93\%) as the concept shift increases from moderate to high.
Similar trend is observed across $\Delta_{Acc}$
as shown in Appendix Figure \ref{fig:appendix_results_sim_maximum}(c).
Thus, as concept shift increases, $\Delta_{Acc}^{group}$ and $\Delta_{Acc}$ increase for NN$_{\not \mathbf{A}}$ considerably in comparison to NN$_{A}$, LR$_{\not \mathbf{A}}$, and LR$_{A}$.
As concept shift increases, the difference in black box model performance between the advantaged and disadvantaged groups for LR$_{A}$, LR$_{\not \mathbf{A}}$, NN$_{A}$, NN$_{\not \mathbf{A}}$ also increases as illustrated in Appendix Figure \ref{fig:bb_sim_1}(c).

\noindent \textbf{Objective 4: Variation in the direct effect of the omitted variable on the outcome}\\
\noindent  As the direct effect of the omitted variable, $C$, on the outcome increases (from 0 to 1.5), excluding $C$ from model training increases disparities in $\Delta_{Acc}^{group}$ when compared to models that include $C$ in model training as shown in Figure \ref{fig:results_sim}(d).
For a direct effect of magnitude 0.5, $\Delta_{Acc}^{group}$ for NN that includes $C$ in model training, represented as NN$_{C}$, is 0.13\% while that of NN excluding $C$, represented as NN$_{\not C}$ is 0.27\%.
While for LR that includes $C$ in model training, LR$_{C}$, $\Delta_{Acc}^{group}$ is 0.05\% for a direct effect of magnitude 0.5, and for LR that excludes $C$ in model training, LR$_{\not C}$, $\Delta_{Acc}^{group}$ is 0.03.
On the contrary, for a direct effect of magnitude 1.5, $\Delta_{Acc}^{group}$ for NN$_{C}$, is 0.09\% while that of NN$_{\not C}$ is 1.55\%, a decrease of 0.04\% and an increase of 1.28\% in comparison to a direct effect of magnitude 0.5, respectively.
While for LR$_{C}$, $\Delta_{Acc}^{group}$ is 0.07\% for a direct effect of magnitude 1.5, and for LR$_{\not C}$, $\Delta_{Acc}^{group}$ is 0.06, an increase of 0.02\% and 0.03\% from a direct effect of magnitude 0.5, respectively.
Thus, $\Delta_{Acc}^{group}$ for NN$_{\not C}$ increases as the direct effect of $C$ on $Y$ increases.
This characteristic is also observed across $\Delta_{Acc}$
as shown in Appendix Figure \ref{fig:appendix_results_sim_maximum}(d).
The difference in black box model performance of  NN$_{\not C}$ is higher than that of  NN$_{C}$, when the direct effect of $C$ on $Y$ is non-zero, precisely for values, 0.5,1.0, and 1.5, as shown in Figure \ref{fig:bb_sim_1}(d).

\subsection{Explanation Disparities in Real-World Dataset}
In the case of the Adult dataset, increasing the percentage of the disadvantaged group (males) in the training sample shows a decrease in $\Delta_{Acc}^{group}$ for LR$_{\not \mathbf{A}}$ and LR$_{A}$ while it remains consistent for NN$_{A}$ and shows a very slight decrease in case of NN$_{\not \mathbf{A}}$. 
This is illustrated in Appendix Figure \ref{fig:appendix_results_adult_1}(b). Similar characteristics can be seen for $\Delta_{Acc}$ in Appendix Figure \ref{fig:appendix_results_adult_1}(a).
Specifically, for 5\% of the disadvantaged group in the training sample,  $\Delta_{Acc}^{group}$ for LR$_{\not \mathbf{A}}$ and LR$_{A}$ is 3.77\% and 2.11\% respectively that decreases to 1.40\% and 1.13\% for 50\% of the disadvantaged group in the training sample. This corresponds to a decrease of 2.37\% for LR$_{A}$ and 0.98 for LR$_{\not \mathbf{A}}$.
While for NN$_{A}$, this decrease corresponds to 0.03\%, and for NN$_{\not \mathbf{A}}$, it is 0.91\% as the percentage of the disadvantaged group (males) increases from 5\% to 50\%.
This differs from the simulation, where explanation disparities in $\Delta_{Acc}^{group}$ remain consistent for LR$_{\not \mathbf{A}}$ and NN$_{\not \mathbf{A}}$ decrease for LR$_{A}$ and NN$_{A}$ whereas in case of Adult, $\Delta_{Acc}^{group}$ for LR$_{\not \mathbf{A}}$, LR$_{A}$, and NN$_{\not \mathbf{A}}$ decrease. 

\begin{table}[!htbp]
\parbox{0.45\textwidth}{
 \centering
    \begin{tabular}{ccc}
    \toprule
        Model & $\Delta_{Acc}$ & $\Delta_{\text{Accuracy}}^{group}$  \\
        \midrule
        LR$_{A}$ & 0.021 (0.020, 0.021)
 & 0.063 (0.062, 0.064)
 \\
        LR$_{\not \mathbf{A}}$ & 0.015 (0.015, 0.015)
  & 0.046 (0.045, 0.047)
\\
        NN$_{A}$& 0.025 (0.025, 0.026)
 & 0.078 (0.077, 0.079)
\\
        NN$_{\not \mathbf{A}}$& 0.022 (0.022, 0.022)
 & 0.067 (0.066, 0.068)
\\
\bottomrule
    \end{tabular}
    \caption{Maximum Fidelity Gap ($\Delta_{Acc}$) and Mean Fidelity Gap in Accuracy ($\Delta_{Acc}^{group}$) with (95\% Confidence interval) for Adult dataset for LR$_{A}$, LR$_{\not \mathbf{A}}$, NN$_{A}$, NN$_{\not \mathbf{A}}$ for concept shift between `hours-per-week' and `income' for male group for Adult (objective 3).}
    \label{tab:adult_3}
}
\hfill
\parbox{0.45\textwidth}{
\centering
    \begin{tabular}{ccc}
    \toprule
      Model   & $\Delta_{Acc}$  & $\Delta_{\text{Accuracy}}^{group}$ \\
    \midrule
        LR$_{C}$ & 0.006 (0.005, 0.006) & 0.017 (0.014, 0.020)\\
        LR$_{\not C}$ & 0.004 (0.004, 0.005)
 & 0.013 (0.011, 0.015)\\
        NN$_{C}$  &  0.027 (0.027, 0.028) & 0.083 (0.081, 0.084) \\
        NN$_{\not C}$&  0.023 (0.022, 0.024) &0.070 (0.068, 0.072) \\
    \bottomrule
    \end{tabular}
    \caption{Maximum Fidelity Gap ($\Delta_{Acc}$) and Mean Fidelity Gap in Accuracy ($\Delta_{Acc}^{group}$) with (95\% Confidence interval) for Adult dataset  for LR with `Nationality' included LR$_C$, LR with `Nationality' excluded LR$_{\not C}$, NN with `Nationality' included NN$_C$ and NN with `Nationality' excluded NN$_{\not C}$ for Adult (objective 4).}
    \label{tab:adult_4}
}
\end{table}

Introducing a covariate shift in $L$ `hours-per-week,' for the disadvantaged group (males), results in the training distribution of the disadvantaged group not being representative of the test distribution of the disadvantaged group.
$\Delta_{Acc}^{group}$ increases from 0.46\% to 0.53\% for LR$_{A}$, 0.50\% to 2.93\% for NN$_{A}$ as percent overlap increases from 20 to 100.
On the contrary, $\Delta_{Acc}^{group}$ for LR$_{\not \mathbf{A}}$ decreases from 2.14\% to 0.41\%, and for NN$_{\not \mathbf{A}}$ it decreases from 4.61\% to 2.94\% as the percent overlap increases from 20 to 100.
This can be seen in Appendix Figure \ref{fig:appendix_results_adult_2}(b).
This differs from simulation, where increasing overlap decreases $\Delta_{Acc}^{group}$ for LR$_{A}$ NN$_{A}$ but $\Delta_{Acc}^{group}$ remains consistent for LR$_{\not \mathbf{A}}$ and NN$_{\not \mathbf{A}}$  while in case of Adult, $\Delta_{Acc}^{group}$ increases for LR$_{A}$, NN$_{A}$ but decreases for LR$_{\not \mathbf{A}}$ and NN$_{\not \mathbf{A}}$.
In the case of concept shift, excluding the sensitive attribute, `gender,' $A$ for model training results in a 1.5\% explanation disparity $\Delta_{Acc}^{group}$ for LR$_{\not \mathbf{A}}$  and 2.2\%  $\Delta_{Acc}^{group}$ for NN$_{\not \mathbf{A}}$. 
Including the sensitive attribute results in an explanation disparity of 2.1\% for LR$_{A}$ (an increase of 0.6\% from LR$_{\not \mathbf{A}}$ and 2.5\% for NN$_{A}$ (an increase of 0.3\% for NN$_{A}$). This is illustrated in Table \ref{tab:adult_3}.
In the case of simulations, excluding $A$ results in higher disparities for LR$_{\not \mathbf{A}}$and NN$_{\not \mathbf{A}}$ in comparison to including $A$, an opposite trend compared to Adult.
Regarding omitted variable bias, excluding `Nationality’ $C$, which has a direct effect on `income’ $Y$, results in an explanation disparity in $\Delta_{Acc}^{group}$ of 2.29\% in comparison to including $C$ in model training, with an explanation disparity of 2.7\% (a difference of 0.49\%) for NN.  While for LR excluding `Nationality' results in 0.52\% explanation disparity, $\Delta_{Acc}^{group}$ in comparison to including it with an explanation disparity, $\Delta_{Acc}^{group}$ of 0.55\% (lower by 0.03\%). This result is illustrated in Table \ref{tab:adult_4}.
Excluding $C$ results in lower disparities for Adult but higher disparities in the case of simulation. Specifically, in the simulations, excluding $C$ results in an explanation disparity in $\Delta_{Acc}^{group}$ of 1.55\% in comparison to including it 0.09\% (a decrease of 1.46\%)  for NN and 0.06\% in explanation disparity in $\Delta_{Acc}^{group}$ for excluding in LR in comparison to 0.06\% while including $C$. 
Since we cannot vary concept shift and the direct effect of `Nationality' on the outcome `Income,' as opposed to the variations in the case of simulations, we only report one value for $\Delta_{Acc}^{group}$, and $\Delta_{Acc}$ for all the models for Adult in Tables \ref{tab:adult_3} and \ref{tab:adult_4} but present multiple values for the simulations in Figures \ref{fig:results_sim}(c) and \ref{fig:results_sim}(d) for objectives 3 and 4, respectively.
Black box model performance for Adult is presented in Appendix Figures  \ref{fig:bb_adut_1}(a) and \ref{fig:bb_adut_1}(b) for objectives 1 and 2, and Appendix Tables \ref{tab:adult_bb_3} and \ref{tab:adult_bb_4} for objectives 3 and 4. These results show that there is a larger difference in the accuracy of black box model performance between groups of the sensitive attribute (males and females) for Adult, 13\% for NN, and 13.8\% for LR in comparison to the simulated data, with 0.10\% for NN and 0.18\% for LR.

\section{Discussion}\label{sec:discussion}
Our study is the first to examine disparities in the LIME explanation method, focusing on the properties of the data and black box model. We examine sample size, covariate shift, concept shift, and omitted variable bias, along with inclusion of the sensitive attribute in the black box model training and the complexity of the black box model.
Our findings show that explanation disparities for both explanation fidelity metrics tested: Maximum Fidelity Gap in accuracy $\Delta_{Acc}$ and Mean Fidelity Gap in accuracy $\Delta_{Acc}^{group}$ can be based on the characteristics of the data as well as the modeling method.
In systematic simulation results, we found no change in $\Delta_{Acc}$ and $\Delta_{Acc}^{group}$ with increasing disadvantaged samples for models that exclude the sensitive attribute if the exclusion of the sensitive attribute aligns with the causal graph.
If the inclusion of the sensitive attribute does not align with the causal graph, $\Delta_{Acc}$ and $\Delta_{Acc}^{group}$ may depend on the proportion of the disadvantaged group in the training sample. 
Here, the black box model is likely to learn spurious correlations between the sensitive attribute and outcome, which can affect $\Delta_{Acc}$ and $\Delta_{Acc}^{group}$.
For limited overlap in the distribution of the disadvantaged sample between the training and test distributions, explanation disparities are higher compared to complete overlap. With a lower overlap, the black box model may not generalize well for the disadvantaged group in the test distribution, resulting in higher disparities in model explanations for $\Delta_{Acc}$ and $\Delta_{Acc}^{group}$.
Moreover, as the concept shift increases, resulting in a nonlinear relationship between the sensitive attribute and the outcome, linear models that are unable to capture this non-linearity actually have lower $\Delta_{Acc}$ and $\Delta_{Acc}^{group}$compared to complex neural network models. 
If the sensitive attribute is excluded from the training of the complex black-box model, not aligning with the causal structure, $\Delta_{Acc}$ and $\Delta_{Acc}^{group}$ are higher than if the sensitive attribute is included in the model training for concept shift. Excluding the sensitive attribute may mask its importance in the predictions of the black box model, resulting in higher explanation disparities.
Omitting a variable that has a direct effect on the outcome may lead to a higher $\Delta_{Acc}$ and $\Delta_{Acc}^{group}$ as the magnitude of the direct effect increases, especially for complex models. As the variable has a direct effect on the outcome, excluding it may mask the importance of the variable in predictions of the black box model, resulting in an increase in explanation disparities.

We find that disparities in model explanations for the Adult dataset reflect the issue of $P(Y|L, C, A) !=  P(Y|L, C)$ in real-world datasets, where it is easier to predict the outcome for the advantaged group rather than the disadvantaged group.
Meanwhile, for the synthetic simulations presented here, it is equally easy to predict across the disadvantaged and advantaged groups.
In general, in simulations, explanation disparities are higher for models that include the sensitive attribute in comparison to models that exclude the sensitive attribute for objectives 1 and 2.
An opposite trend is observed in the case of Adults, where explanation disparities are higher when the sensitive attribute is excluded compared to when it is included.
Including the sensitive attribute aligns with the causal graph for Adult but not for simulations which may explain the differences in the observed behavior between simulations and Adult.
Specifically, as the simpler linear models are not able to capture the nonlinear relationship in Adult, their overall performance increases with an increased proportion of disadvantaged group samples, resulting in a decrease in $\Delta_{Acc}$ and $\Delta_{Acc}^{group}$, irrespective of whether the sensitive attribute is included or not. This differs from simulations where  $\Delta_{Acc}$ and $\Delta_{Acc}^{group}$ only decrease for the models that include the sensitive attribute.
We posit that as our simulations comprised linear functional forms, explanation disparities had a similar trend between linear and neural network models; however, in the case of Adults, since the functional form is nonlinear, explanation disparities for the linear models follow a different trend than neural networks. 
For concept shift, including the sensitive attribute in the black box model training for Adults has a higher $\Delta_{Acc}$ and $\Delta_{Acc}^{group}$ than excluding it in the case of neural networks for concept shift. On the contrary, in simulations, excluding the sensitive attribute resulted in higher $\Delta_{Acc}$ and $\Delta_{Acc}^{group}$.  Excluding the sensitive attribute may result in inaccurate explanations of the black box predictions, especially for complex models such as neural networks. A similar characteristic is observed in Adult when `Nationality' is excluded to assess omitted variable bias, as excluding it can result in inaccurate explanations, especially for the disadvantaged group similar to the simulations.
For mitigating disparities in the explanations in the case of datasets like Adult, our analyses reinforce the need to focus on improving the quality of the data and ensure that the complexity in the data is adequately captured by the black box model.

\noindent \textbf{Need for benchmark datasets for developing fair explanation methods}\\
Given the importance of data, benchmark datasets for assessing explanation disparity metrics would help but currently do not exist.
For developing such benchmark datasets, knowledge of the causal graph can aid in understanding if including sensitive information is relevant to the task and can also highlight which variables can be omitted in the model training to ensure explanations are accurate, especially those that do not directly affect the outcome. 
Further, systematically designed benchmark datasets that allow for varying complexities in the functional form between the covariates and the outcome will be useful in order to assess the explanation disparities of black box models with varying complexity.
Finally, benchmark datasets can be used to assess how different the test distributions can be from the training distributions to ensure that explanation disparities are within the desired range. For example, in the simulations presented here, around 60\% overlap results in a considerable drop in explanation disparities in comparison to 20\% overlap. These types of examinations can help with demonstrating how well a particular explanation method generalizes based on how much a test distribution overlaps with the training distribution.

\noindent \textbf{Limitations of the study}\\
While we highlight the potential factors in the data-generating process and model training that can result in explanation disparities, our study concentrates on the LIME explanation method.
Although LIME has widespread use, and previous research focuses on disparities in LIME explanations, other explanation methods, such as SHAP, may also exhibit disparity challenges that depend on the properties of the data and the black box model. Future work should extend our investigation to other explanation methods, such as SHAP.
Moreover, LIME explanations have inherent limitations, such as uncertainty in perturbation processes. Additionally, the computational cost of LIME and the selection of hyperparameters like kernel width and regularization parameters are crucial to LIME and can influence explanations.
Prior methods developed for addressing these challenges also need to be audited with respect to data and model properties.
While we restrict to simple causal graphs without unmeasured confounding between the sensitive attribute and the outcome, further efforts to include unmeasured confounding can provide additional insights about the relevance of data properties.
However, in considering the properties of the data and models, this work takes an important initial step towards incorporating broader aspects into the assessment of explanation disparities.


\bibliographystyle{ACM-Reference-Format}
\bibliography{references_facct}


\begin{thebibliography}{70}


\ifx \showCODEN    \undefined \def \showCODEN     #1{\unskip}     \fi
\ifx \showDOI      \undefined \def \showDOI       #1{#1}\fi
\ifx \showISBNx    \undefined \def \showISBNx     #1{\unskip}     \fi
\ifx \showISBNxiii \undefined \def \showISBNxiii  #1{\unskip}     \fi
\ifx \showISSN     \undefined \def \showISSN      #1{\unskip}     \fi
\ifx \showLCCN     \undefined \def \showLCCN      #1{\unskip}     \fi
\ifx \shownote     \undefined \def \shownote      #1{#1}          \fi
\ifx \showarticletitle \undefined \def \showarticletitle #1{#1}   \fi
\ifx \showURL      \undefined \def \showURL       {\relax}        \fi
\providecommand\bibfield[2]{#2}
\providecommand\bibinfo[2]{#2}
\providecommand\natexlab[1]{#1}
\providecommand\showeprint[2][]{arXiv:#2}

\bibitem[Adebayo et~al\mbox{.}(2018)]%
        {adebayo2018sanity}
\bibfield{author}{\bibinfo{person}{Julius Adebayo}, \bibinfo{person}{Justin Gilmer}, \bibinfo{person}{Michael Muelly}, \bibinfo{person}{Ian Goodfellow}, \bibinfo{person}{Moritz Hardt}, {and} \bibinfo{person}{Been Kim}.} \bibinfo{year}{2018}\natexlab{}.
\newblock \showarticletitle{Sanity checks for saliency maps}.
\newblock \bibinfo{journal}{\emph{Advances in neural information processing systems}}  \bibinfo{volume}{31} (\bibinfo{year}{2018}).
\newblock


\bibitem[Adebayo et~al\mbox{.}(2022)]%
        {adebayo2022post}
\bibfield{author}{\bibinfo{person}{Julius Adebayo}, \bibinfo{person}{Michael Muelly}, \bibinfo{person}{Hal Abelson}, {and} \bibinfo{person}{Been Kim}.} \bibinfo{year}{2022}\natexlab{}.
\newblock \showarticletitle{Post hoc explanations may be ineffective for detecting unknown spurious correlation}.
\newblock \bibinfo{journal}{\emph{arXiv preprint arXiv:2212.04629}} (\bibinfo{year}{2022}).
\newblock


\bibitem[Adeshola and Adepoju(2023)]%
        {adeshola2023opportunities}
\bibfield{author}{\bibinfo{person}{Ibrahim Adeshola} {and} \bibinfo{person}{Adeola~Praise Adepoju}.} \bibinfo{year}{2023}\natexlab{}.
\newblock \showarticletitle{The opportunities and challenges of ChatGPT in education}.
\newblock \bibinfo{journal}{\emph{Interactive Learning Environments}} (\bibinfo{year}{2023}), \bibinfo{pages}{1--14}.
\newblock


\bibitem[Ahmad et~al\mbox{.}(2018)]%
        {ahmad2018interpretable}
\bibfield{author}{\bibinfo{person}{Muhammad~Aurangzeb Ahmad}, \bibinfo{person}{Carly Eckert}, {and} \bibinfo{person}{Ankur Teredesai}.} \bibinfo{year}{2018}\natexlab{}.
\newblock \showarticletitle{Interpretable machine learning in healthcare}. In \bibinfo{booktitle}{\emph{Proceedings of the 2018 ACM international conference on bioinformatics, computational biology, and health informatics}}. \bibinfo{pages}{559--560}.
\newblock


\bibitem[Allgaier et~al\mbox{.}(2023)]%
        {allgaier2023does}
\bibfield{author}{\bibinfo{person}{Johannes Allgaier}, \bibinfo{person}{Lena Mulansky}, \bibinfo{person}{Rachel~Lea Draelos}, {and} \bibinfo{person}{R{\"u}diger Pryss}.} \bibinfo{year}{2023}\natexlab{}.
\newblock \showarticletitle{How does the model make predictions? A systematic literature review on the explainability power of machine learning in healthcare}.
\newblock \bibinfo{journal}{\emph{Artificial Intelligence in Medicine}}  \bibinfo{volume}{143} (\bibinfo{year}{2023}), \bibinfo{pages}{102616}.
\newblock


\bibitem[Balagopalan et~al\mbox{.}(2022)]%
        {balagopalan2022road}
\bibfield{author}{\bibinfo{person}{Aparna Balagopalan}, \bibinfo{person}{Haoran Zhang}, \bibinfo{person}{Kimia Hamidieh}, \bibinfo{person}{Thomas Hartvigsen}, \bibinfo{person}{Frank Rudzicz}, {and} \bibinfo{person}{Marzyeh Ghassemi}.} \bibinfo{year}{2022}\natexlab{}.
\newblock \showarticletitle{The road to explainability is paved with bias: Measuring the fairness of explanations}. In \bibinfo{booktitle}{\emph{Proceedings of the 2022 ACM Conference on Fairness, Accountability, and Transparency}}. \bibinfo{pages}{1194--1206}.
\newblock


\bibitem[Barocas et~al\mbox{.}(2023)]%
        {barocas-hardt-narayanan}
\bibfield{author}{\bibinfo{person}{Solon Barocas}, \bibinfo{person}{Moritz Hardt}, {and} \bibinfo{person}{Arvind Narayanan}.} \bibinfo{year}{2023}\natexlab{}.
\newblock \bibinfo{booktitle}{\emph{Fairness and Machine Learning: Limitations and Opportunities}}.
\newblock \bibinfo{publisher}{MIT Press}.
\newblock


\bibitem[Begley et~al\mbox{.}(2020)]%
        {begley2020explainability}
\bibfield{author}{\bibinfo{person}{Tom Begley}, \bibinfo{person}{Tobias Schwedes}, \bibinfo{person}{Christopher Frye}, {and} \bibinfo{person}{Ilya Feige}.} \bibinfo{year}{2020}\natexlab{}.
\newblock \showarticletitle{Explainability for fair machine learning}.
\newblock \bibinfo{journal}{\emph{arXiv preprint arXiv:2010.07389}} (\bibinfo{year}{2020}).
\newblock


\bibitem[Bien and Tibshirani(2009)]%
        {bien2009classification}
\bibfield{author}{\bibinfo{person}{Jacob Bien} {and} \bibinfo{person}{Robert Tibshirani}.} \bibinfo{year}{2009}\natexlab{}.
\newblock \showarticletitle{Classification by set cover: The prototype vector machine}.
\newblock \bibinfo{journal}{\emph{arXiv preprint arXiv:0908.2284}} (\bibinfo{year}{2009}).
\newblock


\bibitem[Burkart and Huber(2021)]%
        {burkart2021survey}
\bibfield{author}{\bibinfo{person}{Nadia Burkart} {and} \bibinfo{person}{Marco~F Huber}.} \bibinfo{year}{2021}\natexlab{}.
\newblock \showarticletitle{A survey on the explainability of supervised machine learning}.
\newblock \bibinfo{journal}{\emph{Journal of Artificial Intelligence Research}}  \bibinfo{volume}{70} (\bibinfo{year}{2021}), \bibinfo{pages}{245--317}.
\newblock


\bibitem[Bussmann et~al\mbox{.}(2021)]%
        {bussmann2021explainable}
\bibfield{author}{\bibinfo{person}{Niklas Bussmann}, \bibinfo{person}{Paolo Giudici}, \bibinfo{person}{Dimitri Marinelli}, {and} \bibinfo{person}{Jochen Papenbrock}.} \bibinfo{year}{2021}\natexlab{}.
\newblock \showarticletitle{Explainable machine learning in credit risk management}.
\newblock \bibinfo{journal}{\emph{Computational Economics}}  \bibinfo{volume}{57} (\bibinfo{year}{2021}), \bibinfo{pages}{203--216}.
\newblock


\bibitem[Caruana et~al\mbox{.}(2015)]%
        {caruana2015intelligible}
\bibfield{author}{\bibinfo{person}{Rich Caruana}, \bibinfo{person}{Yin Lou}, \bibinfo{person}{Johannes Gehrke}, \bibinfo{person}{Paul Koch}, \bibinfo{person}{Marc Sturm}, {and} \bibinfo{person}{Noemie Elhadad}.} \bibinfo{year}{2015}\natexlab{}.
\newblock \showarticletitle{Intelligible models for healthcare: Predicting pneumonia risk and hospital 30-day readmission}. In \bibinfo{booktitle}{\emph{Proceedings of the 21th ACM SIGKDD international conference on knowledge discovery and data mining}}. \bibinfo{pages}{1721--1730}.
\newblock


\bibitem[Chen et~al\mbox{.}(2021)]%
        {chen2021ethical}
\bibfield{author}{\bibinfo{person}{Irene~Y Chen}, \bibinfo{person}{Emma Pierson}, \bibinfo{person}{Sherri Rose}, \bibinfo{person}{Shalmali Joshi}, \bibinfo{person}{Kadija Ferryman}, {and} \bibinfo{person}{Marzyeh Ghassemi}.} \bibinfo{year}{2021}\natexlab{}.
\newblock \showarticletitle{Ethical machine learning in healthcare}.
\newblock \bibinfo{journal}{\emph{Annual review of biomedical data science}}  \bibinfo{volume}{4} (\bibinfo{year}{2021}), \bibinfo{pages}{123--144}.
\newblock


\bibitem[Chen et~al\mbox{.}(2023)]%
        {chen2023algorithmic}
\bibfield{author}{\bibinfo{person}{Richard~J Chen}, \bibinfo{person}{Judy~J Wang}, \bibinfo{person}{Drew~FK Williamson}, \bibinfo{person}{Tiffany~Y Chen}, \bibinfo{person}{Jana Lipkova}, \bibinfo{person}{Ming~Y Lu}, \bibinfo{person}{Sharifa Sahai}, {and} \bibinfo{person}{Faisal Mahmood}.} \bibinfo{year}{2023}\natexlab{}.
\newblock \showarticletitle{Algorithmic fairness in artificial intelligence for medicine and healthcare}.
\newblock \bibinfo{journal}{\emph{Nature biomedical engineering}} \bibinfo{volume}{7}, \bibinfo{number}{6} (\bibinfo{year}{2023}), \bibinfo{pages}{719--742}.
\newblock


\bibitem[Chiappa(2019)]%
        {chiappa2019path}
\bibfield{author}{\bibinfo{person}{Silvia Chiappa}.} \bibinfo{year}{2019}\natexlab{}.
\newblock \showarticletitle{Path-specific counterfactual fairness}. In \bibinfo{booktitle}{\emph{Proceedings of the AAAI conference on artificial intelligence}}, Vol.~\bibinfo{volume}{33}. \bibinfo{pages}{7801--7808}.
\newblock


\bibitem[Craven and Shavlik(1995)]%
        {craven1995extracting}
\bibfield{author}{\bibinfo{person}{Mark Craven} {and} \bibinfo{person}{Jude Shavlik}.} \bibinfo{year}{1995}\natexlab{}.
\newblock \showarticletitle{Extracting tree-structured representations of trained networks}.
\newblock \bibinfo{journal}{\emph{Advances in neural information processing systems}}  \bibinfo{volume}{8} (\bibinfo{year}{1995}).
\newblock


\bibitem[Dai et~al\mbox{.}(2022)]%
        {dai2022fairness}
\bibfield{author}{\bibinfo{person}{Jessica Dai}, \bibinfo{person}{Sohini Upadhyay}, \bibinfo{person}{Ulrich Aivodji}, \bibinfo{person}{Stephen~H Bach}, {and} \bibinfo{person}{Himabindu Lakkaraju}.} \bibinfo{year}{2022}\natexlab{}.
\newblock \showarticletitle{Fairness via explanation quality: Evaluating disparities in the quality of post hoc explanations}. In \bibinfo{booktitle}{\emph{Proceedings of the 2022 AAAI/ACM Conference on AI, Ethics, and Society}}. \bibinfo{pages}{203--214}.
\newblock


\bibitem[Dai et~al\mbox{.}(2021)]%
        {dai2021will}
\bibfield{author}{\bibinfo{person}{Jessica Dai}, \bibinfo{person}{Sohini Upadhyay}, \bibinfo{person}{Stephen~H Bach}, {and} \bibinfo{person}{Himabindu Lakkaraju}.} \bibinfo{year}{2021}\natexlab{}.
\newblock \showarticletitle{What will it take to generate fairness-preserving explanations?}
\newblock \bibinfo{journal}{\emph{arXiv preprint arXiv:2106.13346}} (\bibinfo{year}{2021}).
\newblock


\bibitem[Dietvorst et~al\mbox{.}(2015)]%
        {dietvorst2015algorithm}
\bibfield{author}{\bibinfo{person}{Berkeley~J Dietvorst}, \bibinfo{person}{Joseph~P Simmons}, {and} \bibinfo{person}{Cade Massey}.} \bibinfo{year}{2015}\natexlab{}.
\newblock \showarticletitle{Algorithm aversion: people erroneously avoid algorithms after seeing them err.}
\newblock \bibinfo{journal}{\emph{Journal of Experimental Psychology: General}} \bibinfo{volume}{144}, \bibinfo{number}{1} (\bibinfo{year}{2015}), \bibinfo{pages}{114}.
\newblock


\bibitem[Do{\v{s}}ilovi{\'c} et~al\mbox{.}(2018)]%
        {dovsilovic2018explainable}
\bibfield{author}{\bibinfo{person}{Filip~Karlo Do{\v{s}}ilovi{\'c}}, \bibinfo{person}{Mario Br{\v{c}}i{\'c}}, {and} \bibinfo{person}{Nikica Hlupi{\'c}}.} \bibinfo{year}{2018}\natexlab{}.
\newblock \showarticletitle{Explainable artificial intelligence: A survey}. In \bibinfo{booktitle}{\emph{2018 41st International convention on information and communication technology, electronics and microelectronics (MIPRO)}}. IEEE, \bibinfo{pages}{0210--0215}.
\newblock


\bibitem[Dressel and Farid(2018)]%
        {dressel2018accuracy}
\bibfield{author}{\bibinfo{person}{Julia Dressel} {and} \bibinfo{person}{Hany Farid}.} \bibinfo{year}{2018}\natexlab{}.
\newblock \showarticletitle{The accuracy, fairness, and limits of predicting recidivism}.
\newblock \bibinfo{journal}{\emph{Science advances}} \bibinfo{volume}{4}, \bibinfo{number}{1} (\bibinfo{year}{2018}), \bibinfo{pages}{eaao5580}.
\newblock


\bibitem[Dua et~al\mbox{.}(2017)]%
        {dua2017uci}
\bibfield{author}{\bibinfo{person}{Dheeru Dua}, \bibinfo{person}{Casey Graff}, {et~al\mbox{.}}} \bibinfo{year}{2017}\natexlab{}.
\newblock \showarticletitle{UCI machine learning repository}.
\newblock  (\bibinfo{year}{2017}).
\newblock


\bibitem[Fabris et~al\mbox{.}(2022)]%
        {fabris2022algorithmic}
\bibfield{author}{\bibinfo{person}{Alessandro Fabris}, \bibinfo{person}{Stefano Messina}, \bibinfo{person}{Gianmaria Silvello}, {and} \bibinfo{person}{Gian~Antonio Susto}.} \bibinfo{year}{2022}\natexlab{}.
\newblock \showarticletitle{Algorithmic fairness datasets: the story so far}.
\newblock \bibinfo{journal}{\emph{Data Mining and Knowledge Discovery}} \bibinfo{volume}{36}, \bibinfo{number}{6} (\bibinfo{year}{2022}), \bibinfo{pages}{2074--2152}.
\newblock


\bibitem[Gebru et~al\mbox{.}(2021)]%
        {gebru2021datasheets}
\bibfield{author}{\bibinfo{person}{Timnit Gebru}, \bibinfo{person}{Jamie Morgenstern}, \bibinfo{person}{Briana Vecchione}, \bibinfo{person}{Jennifer~Wortman Vaughan}, \bibinfo{person}{Hanna Wallach}, \bibinfo{person}{Hal~Daum{\'e} Iii}, {and} \bibinfo{person}{Kate Crawford}.} \bibinfo{year}{2021}\natexlab{}.
\newblock \showarticletitle{Datasheets for datasets}.
\newblock \bibinfo{journal}{\emph{Commun. ACM}} \bibinfo{volume}{64}, \bibinfo{number}{12} (\bibinfo{year}{2021}), \bibinfo{pages}{86--92}.
\newblock


\bibitem[Ghassemi et~al\mbox{.}(2020)]%
        {ghassemi2020review}
\bibfield{author}{\bibinfo{person}{Marzyeh Ghassemi}, \bibinfo{person}{Tristan Naumann}, \bibinfo{person}{Peter Schulam}, \bibinfo{person}{Andrew~L Beam}, \bibinfo{person}{Irene~Y Chen}, {and} \bibinfo{person}{Rajesh Ranganath}.} \bibinfo{year}{2020}\natexlab{}.
\newblock \showarticletitle{A review of challenges and opportunities in machine learning for health}.
\newblock \bibinfo{journal}{\emph{AMIA Summits on Translational Science Proceedings}}  \bibinfo{volume}{2020} (\bibinfo{year}{2020}), \bibinfo{pages}{191}.
\newblock


\bibitem[Hase and Bansal(2020)]%
        {hase2020evaluating}
\bibfield{author}{\bibinfo{person}{Peter Hase} {and} \bibinfo{person}{Mohit Bansal}.} \bibinfo{year}{2020}\natexlab{}.
\newblock \showarticletitle{Evaluating explainable AI: Which algorithmic explanations help users predict model behavior?}
\newblock \bibinfo{journal}{\emph{arXiv preprint arXiv:2005.01831}} (\bibinfo{year}{2020}).
\newblock


\bibitem[Kallus and Zhou(2018)]%
        {kallus2018residual}
\bibfield{author}{\bibinfo{person}{Nathan Kallus} {and} \bibinfo{person}{Angela Zhou}.} \bibinfo{year}{2018}\natexlab{}.
\newblock \showarticletitle{Residual unfairness in fair machine learning from prejudiced data}. In \bibinfo{booktitle}{\emph{International Conference on Machine Learning}}. PMLR, \bibinfo{pages}{2439--2448}.
\newblock


\bibitem[Kamishima et~al\mbox{.}(2012)]%
        {kamishima2012fairness}
\bibfield{author}{\bibinfo{person}{Toshihiro Kamishima}, \bibinfo{person}{Shotaro Akaho}, \bibinfo{person}{Hideki Asoh}, {and} \bibinfo{person}{Jun Sakuma}.} \bibinfo{year}{2012}\natexlab{}.
\newblock \showarticletitle{Fairness-aware classifier with prejudice remover regularizer}. In \bibinfo{booktitle}{\emph{Machine Learning and Knowledge Discovery in Databases: European Conference, ECML PKDD 2012, Bristol, UK, September 24-28, 2012. Proceedings, Part II 23}}. Springer, \bibinfo{pages}{35--50}.
\newblock


\bibitem[Karimi et~al\mbox{.}(2020)]%
        {karimi2020model}
\bibfield{author}{\bibinfo{person}{Amir-Hossein Karimi}, \bibinfo{person}{Gilles Barthe}, \bibinfo{person}{Borja Balle}, {and} \bibinfo{person}{Isabel Valera}.} \bibinfo{year}{2020}\natexlab{}.
\newblock \showarticletitle{Model-agnostic counterfactual explanations for consequential decisions}. In \bibinfo{booktitle}{\emph{International Conference on Artificial Intelligence and Statistics}}. PMLR, \bibinfo{pages}{895--905}.
\newblock


\bibitem[Kleinberg et~al\mbox{.}(2022)]%
        {kleinberg2022racial}
\bibfield{author}{\bibinfo{person}{Giona Kleinberg}, \bibinfo{person}{Michael~J Diaz}, \bibinfo{person}{Sai Batchu}, {and} \bibinfo{person}{Brandon Lucke-Wold}.} \bibinfo{year}{2022}\natexlab{}.
\newblock \showarticletitle{Racial underrepresentation in dermatological datasets leads to biased machine learning models and inequitable healthcare}.
\newblock \bibinfo{journal}{\emph{Journal of biomed research}} \bibinfo{volume}{3}, \bibinfo{number}{1} (\bibinfo{year}{2022}), \bibinfo{pages}{42}.
\newblock


\bibitem[Lakkaraju et~al\mbox{.}(2016)]%
        {lakkaraju2016interpretable}
\bibfield{author}{\bibinfo{person}{Himabindu Lakkaraju}, \bibinfo{person}{Stephen~H Bach}, {and} \bibinfo{person}{Jure Leskovec}.} \bibinfo{year}{2016}\natexlab{}.
\newblock \showarticletitle{Interpretable decision sets: A joint framework for description and prediction}. In \bibinfo{booktitle}{\emph{Proceedings of the 22nd ACM SIGKDD international conference on knowledge discovery and data mining}}. \bibinfo{pages}{1675--1684}.
\newblock


\bibitem[Laugel et~al\mbox{.}(2019)]%
        {laugel2019dangers}
\bibfield{author}{\bibinfo{person}{Thibault Laugel}, \bibinfo{person}{Marie-Jeanne Lesot}, \bibinfo{person}{Christophe Marsala}, \bibinfo{person}{Xavier Renard}, {and} \bibinfo{person}{Marcin Detyniecki}.} \bibinfo{year}{2019}\natexlab{}.
\newblock \showarticletitle{The dangers of post-hoc interpretability: Unjustified counterfactual explanations}.
\newblock \bibinfo{journal}{\emph{arXiv preprint arXiv:1907.09294}} (\bibinfo{year}{2019}).
\newblock


\bibitem[Letham et~al\mbox{.}(2015)]%
        {letham2015interpretable}
\bibfield{author}{\bibinfo{person}{Benjamin Letham}, \bibinfo{person}{Cynthia Rudin}, \bibinfo{person}{Tyler~H McCormick}, {and} \bibinfo{person}{David Madigan}.} \bibinfo{year}{2015}\natexlab{}.
\newblock \showarticletitle{Interpretable classifiers using rules and bayesian analysis: Building a better stroke prediction model}.
\newblock  (\bibinfo{year}{2015}).
\newblock


\bibitem[Lichman et~al\mbox{.}(2013)]%
        {lichman2013uci}
\bibfield{author}{\bibinfo{person}{Moshe Lichman} {et~al\mbox{.}}} \bibinfo{year}{2013}\natexlab{}.
\newblock \bibinfo{title}{UCI machine learning repository}.
\newblock
\newblock


\bibitem[Linardatos et~al\mbox{.}(2020)]%
        {linardatos2020explainable}
\bibfield{author}{\bibinfo{person}{Pantelis Linardatos}, \bibinfo{person}{Vasilis Papastefanopoulos}, {and} \bibinfo{person}{Sotiris Kotsiantis}.} \bibinfo{year}{2020}\natexlab{}.
\newblock \showarticletitle{Explainable ai: A review of machine learning interpretability methods}.
\newblock \bibinfo{journal}{\emph{Entropy}} \bibinfo{volume}{23}, \bibinfo{number}{1} (\bibinfo{year}{2020}), \bibinfo{pages}{18}.
\newblock


\bibitem[Liu et~al\mbox{.}(2021)]%
        {liu2021just}
\bibfield{author}{\bibinfo{person}{Evan~Z Liu}, \bibinfo{person}{Behzad Haghgoo}, \bibinfo{person}{Annie~S Chen}, \bibinfo{person}{Aditi Raghunathan}, \bibinfo{person}{Pang~Wei Koh}, \bibinfo{person}{Shiori Sagawa}, \bibinfo{person}{Percy Liang}, {and} \bibinfo{person}{Chelsea Finn}.} \bibinfo{year}{2021}\natexlab{}.
\newblock \showarticletitle{Just train twice: Improving group robustness without training group information}. In \bibinfo{booktitle}{\emph{International Conference on Machine Learning}}. PMLR, \bibinfo{pages}{6781--6792}.
\newblock


\bibitem[Lundberg and Lee(2017)]%
        {lundberg2017unified}
\bibfield{author}{\bibinfo{person}{Scott~M Lundberg} {and} \bibinfo{person}{Su-In Lee}.} \bibinfo{year}{2017}\natexlab{}.
\newblock \showarticletitle{A unified approach to interpreting model predictions}.
\newblock \bibinfo{journal}{\emph{Advances in neural information processing systems}}  \bibinfo{volume}{30} (\bibinfo{year}{2017}).
\newblock


\bibitem[Mart{\'\i}nez-Plumed et~al\mbox{.}(2019)]%
        {martinez2019fairness}
\bibfield{author}{\bibinfo{person}{Fernando Mart{\'\i}nez-Plumed}, \bibinfo{person}{C{\`e}sar Ferri}, \bibinfo{person}{David Nieves}, {and} \bibinfo{person}{Jos{\'e} Hern{\'a}ndez-Orallo}.} \bibinfo{year}{2019}\natexlab{}.
\newblock \showarticletitle{Fairness and missing values}.
\newblock \bibinfo{journal}{\emph{arXiv preprint arXiv:1905.12728}} (\bibinfo{year}{2019}).
\newblock


\bibitem[Mhasawade and Chunara(2021)]%
        {mhasawade2021causal}
\bibfield{author}{\bibinfo{person}{Vishwali Mhasawade} {and} \bibinfo{person}{Rumi Chunara}.} \bibinfo{year}{2021}\natexlab{}.
\newblock \showarticletitle{Causal multi-level fairness}. In \bibinfo{booktitle}{\emph{Proceedings of the 2021 AAAI/ACM Conference on AI, Ethics, and Society}}. \bibinfo{pages}{784--794}.
\newblock


\bibitem[Mitchell et~al\mbox{.}(2019)]%
        {mitchell2019model}
\bibfield{author}{\bibinfo{person}{Margaret Mitchell}, \bibinfo{person}{Simone Wu}, \bibinfo{person}{Andrew Zaldivar}, \bibinfo{person}{Parker Barnes}, \bibinfo{person}{Lucy Vasserman}, \bibinfo{person}{Ben Hutchinson}, \bibinfo{person}{Elena Spitzer}, \bibinfo{person}{Inioluwa~Deborah Raji}, {and} \bibinfo{person}{Timnit Gebru}.} \bibinfo{year}{2019}\natexlab{}.
\newblock \showarticletitle{Model cards for model reporting}. In \bibinfo{booktitle}{\emph{Proceedings of the conference on fairness, accountability, and transparency}}. \bibinfo{pages}{220--229}.
\newblock


\bibitem[Moreno-Torres et~al\mbox{.}(2012)]%
        {moreno2012unifying}
\bibfield{author}{\bibinfo{person}{Jose~G Moreno-Torres}, \bibinfo{person}{Troy Raeder}, \bibinfo{person}{Roc{\'\i}o Alaiz-Rodr{\'\i}guez}, \bibinfo{person}{Nitesh~V Chawla}, {and} \bibinfo{person}{Francisco Herrera}.} \bibinfo{year}{2012}\natexlab{}.
\newblock \showarticletitle{A unifying view on dataset shift in classification}.
\newblock \bibinfo{journal}{\emph{Pattern recognition}} \bibinfo{volume}{45}, \bibinfo{number}{1} (\bibinfo{year}{2012}), \bibinfo{pages}{521--530}.
\newblock


\bibitem[Namkoong et~al\mbox{.}(2023)]%
        {namkoong2023diagnosing}
\bibfield{author}{\bibinfo{person}{Hongseok Namkoong}, \bibinfo{person}{Steve Yadlowsky}, {et~al\mbox{.}}} \bibinfo{year}{2023}\natexlab{}.
\newblock \showarticletitle{Diagnosing Model Performance Under Distribution Shift}.
\newblock \bibinfo{journal}{\emph{arXiv preprint arXiv:2303.02011}} (\bibinfo{year}{2023}).
\newblock


\bibitem[Pessach and Shmueli(2023)]%
        {pessach2023algorithmic}
\bibfield{author}{\bibinfo{person}{Dana Pessach} {and} \bibinfo{person}{Erez Shmueli}.} \bibinfo{year}{2023}\natexlab{}.
\newblock \showarticletitle{Algorithmic fairness}.
\newblock In \bibinfo{booktitle}{\emph{Machine Learning for Data Science Handbook: Data Mining and Knowledge Discovery Handbook}}. \bibinfo{publisher}{Springer}, \bibinfo{pages}{867--886}.
\newblock


\bibitem[Pfohl et~al\mbox{.}(2023)]%
        {pfohl2023understanding}
\bibfield{author}{\bibinfo{person}{Stephen~Robert Pfohl}, \bibinfo{person}{Natalie Harris}, \bibinfo{person}{Chirag Nagpal}, \bibinfo{person}{David Madras}, \bibinfo{person}{Vishwali Mhasawade}, \bibinfo{person}{Olawale~Elijah Salaudeen}, \bibinfo{person}{Katherine~A Heller}, \bibinfo{person}{Sanmi Koyejo}, {and} \bibinfo{person}{Alexander~Nicholas D'Amour}.} \bibinfo{year}{2023}\natexlab{}.
\newblock \showarticletitle{Understanding subgroup performance differences of fair predictors using causal models}. In \bibinfo{booktitle}{\emph{NeurIPS 2023 Workshop on Distribution Shifts: New Frontiers with Foundation Models}}.
\newblock


\bibitem[Plumb et~al\mbox{.}(2018)]%
        {plumb2018model}
\bibfield{author}{\bibinfo{person}{Gregory Plumb}, \bibinfo{person}{Denali Molitor}, {and} \bibinfo{person}{Ameet~S Talwalkar}.} \bibinfo{year}{2018}\natexlab{}.
\newblock \showarticletitle{Model agnostic supervised local explanations}.
\newblock \bibinfo{journal}{\emph{Advances in neural information processing systems}}  \bibinfo{volume}{31} (\bibinfo{year}{2018}).
\newblock


\bibitem[Ribeiro et~al\mbox{.}(2016)]%
        {ribeiro2016should}
\bibfield{author}{\bibinfo{person}{Marco~Tulio Ribeiro}, \bibinfo{person}{Sameer Singh}, {and} \bibinfo{person}{Carlos Guestrin}.} \bibinfo{year}{2016}\natexlab{}.
\newblock \showarticletitle{" Why should i trust you?" Explaining the predictions of any classifier}. In \bibinfo{booktitle}{\emph{Proceedings of the 22nd ACM SIGKDD international conference on knowledge discovery and data mining}}. \bibinfo{pages}{1135--1144}.
\newblock


\bibitem[Ribeiro et~al\mbox{.}(2018)]%
        {ribeiro2018anchors}
\bibfield{author}{\bibinfo{person}{Marco~Tulio Ribeiro}, \bibinfo{person}{Sameer Singh}, {and} \bibinfo{person}{Carlos Guestrin}.} \bibinfo{year}{2018}\natexlab{}.
\newblock \showarticletitle{Anchors: High-precision model-agnostic explanations}. In \bibinfo{booktitle}{\emph{Proceedings of the AAAI conference on artificial intelligence}}, Vol.~\bibinfo{volume}{32}.
\newblock


\bibitem[Ricci~Lara et~al\mbox{.}(2023)]%
        {ricci2023towards}
\bibfield{author}{\bibinfo{person}{Mar{\'\i}a~Agustina Ricci~Lara}, \bibinfo{person}{Candelaria Mosquera}, \bibinfo{person}{Enzo Ferrante}, {and} \bibinfo{person}{Rodrigo Echeveste}.} \bibinfo{year}{2023}\natexlab{}.
\newblock \showarticletitle{Towards Unraveling Calibration Biases in Medical Image Analysis}. In \bibinfo{booktitle}{\emph{Workshop on Clinical Image-Based Procedures}}. Springer, \bibinfo{pages}{132--141}.
\newblock


\bibitem[Rudin(2019)]%
        {rudin2019stop}
\bibfield{author}{\bibinfo{person}{Cynthia Rudin}.} \bibinfo{year}{2019}\natexlab{}.
\newblock \showarticletitle{Stop explaining black box machine learning models for high stakes decisions and use interpretable models instead}.
\newblock \bibinfo{journal}{\emph{Nature machine intelligence}} \bibinfo{volume}{1}, \bibinfo{number}{5} (\bibinfo{year}{2019}), \bibinfo{pages}{206--215}.
\newblock


\bibitem[Sangroya et~al\mbox{.}(2020)]%
        {sangroya2020guided}
\bibfield{author}{\bibinfo{person}{Amit Sangroya}, \bibinfo{person}{Mouli Rastogi}, \bibinfo{person}{C Anantaram}, {and} \bibinfo{person}{Lovekesh Vig}.} \bibinfo{year}{2020}\natexlab{}.
\newblock \showarticletitle{Guided-LIME: Structured Sampling based Hybrid Approach towards Explaining Blackbox Machine Learning Models.}. In \bibinfo{booktitle}{\emph{CIKM (Workshops)}}.
\newblock


\bibitem[Selvaraju et~al\mbox{.}(2017)]%
        {selvaraju2017grad}
\bibfield{author}{\bibinfo{person}{Ramprasaath~R Selvaraju}, \bibinfo{person}{Michael Cogswell}, \bibinfo{person}{Abhishek Das}, \bibinfo{person}{Ramakrishna Vedantam}, \bibinfo{person}{Devi Parikh}, {and} \bibinfo{person}{Dhruv Batra}.} \bibinfo{year}{2017}\natexlab{}.
\newblock \showarticletitle{Grad-cam: Visual explanations from deep networks via gradient-based localization}. In \bibinfo{booktitle}{\emph{Proceedings of the IEEE international conference on computer vision}}. \bibinfo{pages}{618--626}.
\newblock


\bibitem[Singh et~al\mbox{.}(2021)]%
        {singh2021fairness}
\bibfield{author}{\bibinfo{person}{Harvineet Singh}, \bibinfo{person}{Rina Singh}, \bibinfo{person}{Vishwali Mhasawade}, {and} \bibinfo{person}{Rumi Chunara}.} \bibinfo{year}{2021}\natexlab{}.
\newblock \showarticletitle{Fairness violations and mitigation under covariate shift}. In \bibinfo{booktitle}{\emph{Proceedings of the 2021 ACM Conference on Fairness, Accountability, and Transparency}}. \bibinfo{pages}{3--13}.
\newblock


\bibitem[Slack et~al\mbox{.}(2021)]%
        {slack2021reliable}
\bibfield{author}{\bibinfo{person}{Dylan Slack}, \bibinfo{person}{Anna Hilgard}, \bibinfo{person}{Sameer Singh}, {and} \bibinfo{person}{Himabindu Lakkaraju}.} \bibinfo{year}{2021}\natexlab{}.
\newblock \showarticletitle{Reliable post hoc explanations: Modeling uncertainty in explainability}.
\newblock \bibinfo{journal}{\emph{Advances in neural information processing systems}}  \bibinfo{volume}{34} (\bibinfo{year}{2021}), \bibinfo{pages}{9391--9404}.
\newblock


\bibitem[Smilkov et~al\mbox{.}(2017)]%
        {smilkov2017smoothgrad}
\bibfield{author}{\bibinfo{person}{Daniel Smilkov}, \bibinfo{person}{Nikhil Thorat}, \bibinfo{person}{Been Kim}, \bibinfo{person}{Fernanda Vi{\'e}gas}, {and} \bibinfo{person}{Martin Wattenberg}.} \bibinfo{year}{2017}\natexlab{}.
\newblock \showarticletitle{Smoothgrad: removing noise by adding noise}.
\newblock \bibinfo{journal}{\emph{arXiv preprint arXiv:1706.03825}} (\bibinfo{year}{2017}).
\newblock


\bibitem[Subbaswamy and Saria(2018)]%
        {subbaswamy2018counterfactual}
\bibfield{author}{\bibinfo{person}{Adarsh Subbaswamy} {and} \bibinfo{person}{Suchi Saria}.} \bibinfo{year}{2018}\natexlab{}.
\newblock \showarticletitle{Counterfactual Normalization: Proactively Addressing Dataset Shift Using Causal Mechanisms.}. In \bibinfo{booktitle}{\emph{UAI}}. \bibinfo{pages}{947--957}.
\newblock


\bibitem[Subbaswamy and Saria(2020)]%
        {subbaswamy2020development}
\bibfield{author}{\bibinfo{person}{Adarsh Subbaswamy} {and} \bibinfo{person}{Suchi Saria}.} \bibinfo{year}{2020}\natexlab{}.
\newblock \showarticletitle{From development to deployment: dataset shift, causality, and shift-stable models in health AI}.
\newblock \bibinfo{journal}{\emph{Biostatistics}} \bibinfo{volume}{21}, \bibinfo{number}{2} (\bibinfo{year}{2020}), \bibinfo{pages}{345--352}.
\newblock


\bibitem[Tan et~al\mbox{.}(2018)]%
        {tan2018distill}
\bibfield{author}{\bibinfo{person}{Sarah Tan}, \bibinfo{person}{Rich Caruana}, \bibinfo{person}{Giles Hooker}, {and} \bibinfo{person}{Yin Lou}.} \bibinfo{year}{2018}\natexlab{}.
\newblock \showarticletitle{Distill-and-compare: Auditing black-box models using transparent model distillation}. In \bibinfo{booktitle}{\emph{Proceedings of the 2018 AAAI/ACM Conference on AI, Ethics, and Society}}. \bibinfo{pages}{303--310}.
\newblock


\bibitem[Tuggener et~al\mbox{.}(2019)]%
        {tuggener2019automated}
\bibfield{author}{\bibinfo{person}{Lukas Tuggener}, \bibinfo{person}{Mohammadreza Amirian}, \bibinfo{person}{Katharina Rombach}, \bibinfo{person}{Stefan L{\"o}rwald}, \bibinfo{person}{Anastasia Varlet}, \bibinfo{person}{Christian Westermann}, {and} \bibinfo{person}{Thilo Stadelmann}.} \bibinfo{year}{2019}\natexlab{}.
\newblock \showarticletitle{Automated machine learning in practice: state of the art and recent results}. In \bibinfo{booktitle}{\emph{2019 6th Swiss Conference on Data Science (SDS)}}. IEEE, \bibinfo{pages}{31--36}.
\newblock


\bibitem[Velmurugan et~al\mbox{.}(2021)]%
        {velmurugan2021developing}
\bibfield{author}{\bibinfo{person}{Mythreyi Velmurugan}, \bibinfo{person}{Chun Ouyang}, \bibinfo{person}{Catarina Moreira}, {and} \bibinfo{person}{Renuka Sindhgatta}.} \bibinfo{year}{2021}\natexlab{}.
\newblock \showarticletitle{Developing a fidelity evaluation approach for interpretable machine learning}.
\newblock \bibinfo{journal}{\emph{arXiv preprint arXiv:2106.08492}} (\bibinfo{year}{2021}).
\newblock


\bibitem[Vyas et~al\mbox{.}(2020)]%
        {vyas2020hidden}
\bibfield{author}{\bibinfo{person}{Darshali~A Vyas}, \bibinfo{person}{Leo~G Eisenstein}, {and} \bibinfo{person}{David~S Jones}.} \bibinfo{year}{2020}\natexlab{}.
\newblock \bibinfo{title}{Hidden in plain sight—reconsidering the use of race correction in clinical algorithms}.
\newblock , \bibinfo{numpages}{874--882}~pages.
\newblock


\bibitem[Wachter et~al\mbox{.}(2017)]%
        {wachter2017counterfactual}
\bibfield{author}{\bibinfo{person}{Sandra Wachter}, \bibinfo{person}{Brent Mittelstadt}, {and} \bibinfo{person}{Chris Russell}.} \bibinfo{year}{2017}\natexlab{}.
\newblock \showarticletitle{Counterfactual explanations without opening the black box: Automated decisions and the GDPR}.
\newblock \bibinfo{journal}{\emph{Harv. JL \& Tech.}}  \bibinfo{volume}{31} (\bibinfo{year}{2017}), \bibinfo{pages}{841}.
\newblock


\bibitem[Wang and Rudin(2015)]%
        {wang2015falling}
\bibfield{author}{\bibinfo{person}{Fulton Wang} {and} \bibinfo{person}{Cynthia Rudin}.} \bibinfo{year}{2015}\natexlab{}.
\newblock \showarticletitle{Falling rule lists}. In \bibinfo{booktitle}{\emph{Artificial intelligence and statistics}}. PMLR, \bibinfo{pages}{1013--1022}.
\newblock


\bibitem[Wang and Singh(2021)]%
        {wang2021analyzing}
\bibfield{author}{\bibinfo{person}{Yanchen Wang} {and} \bibinfo{person}{Lisa Singh}.} \bibinfo{year}{2021}\natexlab{}.
\newblock \showarticletitle{Analyzing the impact of missing values and selection bias on fairness}.
\newblock \bibinfo{journal}{\emph{International Journal of Data Science and Analytics}} \bibinfo{volume}{12}, \bibinfo{number}{2} (\bibinfo{year}{2021}), \bibinfo{pages}{101--119}.
\newblock


\bibitem[Wu et~al\mbox{.}(2023)]%
        {wu2023bloomberggpt}
\bibfield{author}{\bibinfo{person}{Shijie Wu}, \bibinfo{person}{Ozan Irsoy}, \bibinfo{person}{Steven Lu}, \bibinfo{person}{Vadim Dabravolski}, \bibinfo{person}{Mark Dredze}, \bibinfo{person}{Sebastian Gehrmann}, \bibinfo{person}{Prabhanjan Kambadur}, \bibinfo{person}{David Rosenberg}, {and} \bibinfo{person}{Gideon Mann}.} \bibinfo{year}{2023}\natexlab{}.
\newblock \showarticletitle{Bloomberggpt: A large language model for finance}.
\newblock \bibinfo{journal}{\emph{arXiv preprint arXiv:2303.17564}} (\bibinfo{year}{2023}).
\newblock


\bibitem[Yosinski et~al\mbox{.}(2015)]%
        {yosinski2015understanding}
\bibfield{author}{\bibinfo{person}{Jason Yosinski}, \bibinfo{person}{Jeff Clune}, \bibinfo{person}{Anh Nguyen}, \bibinfo{person}{Thomas Fuchs}, {and} \bibinfo{person}{Hod Lipson}.} \bibinfo{year}{2015}\natexlab{}.
\newblock \showarticletitle{Understanding neural networks through deep visualization}.
\newblock \bibinfo{journal}{\emph{arXiv preprint arXiv:1506.06579}} (\bibinfo{year}{2015}).
\newblock


\bibitem[Zafar et~al\mbox{.}(2017)]%
        {zafar2017fairness}
\bibfield{author}{\bibinfo{person}{Muhammad~Bilal Zafar}, \bibinfo{person}{Isabel Valera}, \bibinfo{person}{Manuel~Gomez Rogriguez}, {and} \bibinfo{person}{Krishna~P Gummadi}.} \bibinfo{year}{2017}\natexlab{}.
\newblock \showarticletitle{Fairness constraints: Mechanisms for fair classification}. In \bibinfo{booktitle}{\emph{Artificial intelligence and statistics}}. PMLR, \bibinfo{pages}{962--970}.
\newblock


\bibitem[Zeng et~al\mbox{.}(2017)]%
        {zeng2017interpretable}
\bibfield{author}{\bibinfo{person}{Jiaming Zeng}, \bibinfo{person}{Berk Ustun}, {and} \bibinfo{person}{Cynthia Rudin}.} \bibinfo{year}{2017}\natexlab{}.
\newblock \showarticletitle{Interpretable classification models for recidivism prediction}.
\newblock \bibinfo{journal}{\emph{Journal of the Royal Statistical Society Series A: Statistics in Society}} \bibinfo{volume}{180}, \bibinfo{number}{3} (\bibinfo{year}{2017}), \bibinfo{pages}{689--722}.
\newblock


\bibitem[Zhao et~al\mbox{.}(2023)]%
        {zhao2023machine}
\bibfield{author}{\bibinfo{person}{Yijun Zhao}, \bibinfo{person}{Xiaoyu Chen}, \bibinfo{person}{Haoran Xue}, {and} \bibinfo{person}{Gary~M Weiss}.} \bibinfo{year}{2023}\natexlab{}.
\newblock \showarticletitle{A machine learning approach to graduate admissions and the role of letters of recommendation}.
\newblock \bibinfo{journal}{\emph{Plos one}} \bibinfo{volume}{18}, \bibinfo{number}{10} (\bibinfo{year}{2023}), \bibinfo{pages}{e0291107}.
\newblock


\bibitem[Zhou et~al\mbox{.}(2021a)]%
        {zhou2021evaluating}
\bibfield{author}{\bibinfo{person}{Jianlong Zhou}, \bibinfo{person}{Amir~H Gandomi}, \bibinfo{person}{Fang Chen}, {and} \bibinfo{person}{Andreas Holzinger}.} \bibinfo{year}{2021}\natexlab{a}.
\newblock \showarticletitle{Evaluating the quality of machine learning explanations: A survey on methods and metrics}.
\newblock \bibinfo{journal}{\emph{Electronics}} \bibinfo{volume}{10}, \bibinfo{number}{5} (\bibinfo{year}{2021}), \bibinfo{pages}{593}.
\newblock


\bibitem[Zhou et~al\mbox{.}(2021b)]%
        {zhou2021s}
\bibfield{author}{\bibinfo{person}{Zhengze Zhou}, \bibinfo{person}{Giles Hooker}, {and} \bibinfo{person}{Fei Wang}.} \bibinfo{year}{2021}\natexlab{b}.
\newblock \showarticletitle{S-lime: Stabilized-lime for model explanation}. In \bibinfo{booktitle}{\emph{Proceedings of the 27th ACM SIGKDD conference on knowledge discovery \& data mining}}. \bibinfo{pages}{2429--2438}.
\newblock


\end{thebibliography}
\newpage
\appendix
\section{Appendix}
\subsection{Additional Performance Metric Results for Simulation}
Results for $\Delta_{Acc}$ for all 4 objectives are presented in Appendix Figure \ref{fig:appendix_results_sim_maximum}.
We also provide detailed results for the simulation pertaining to the black box model performance for all 4 objectives in Appendix Figure \ref{fig:bb_sim_1}. 

\begin{figure}[h]
    \centering
    \begin{subfigure}[b]{0.45\textwidth}
    \centering
    \includegraphics[scale=0.18]{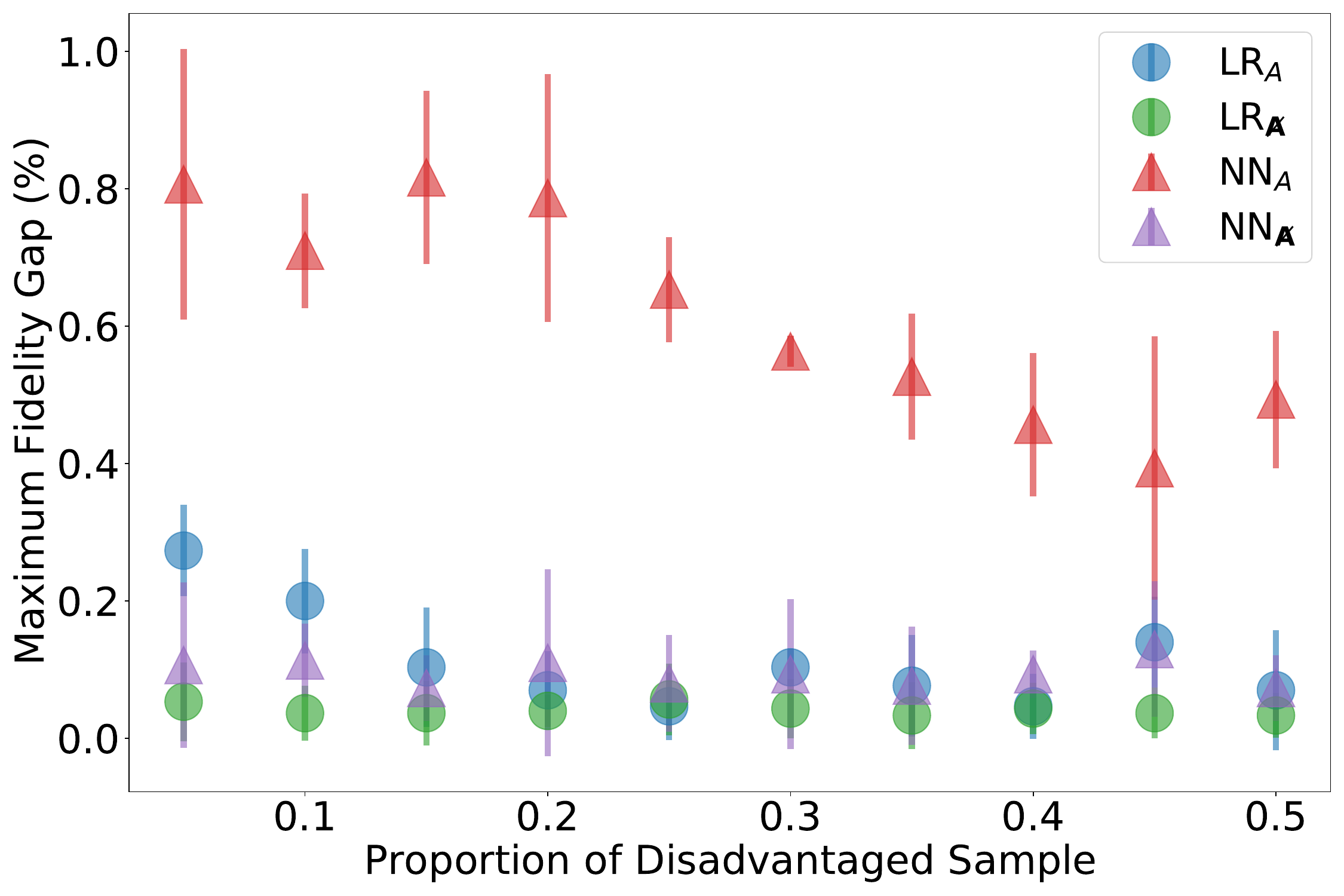}
    \caption{}
    \end{subfigure}
    \hfill
    \begin{subfigure}[b]{0.45\textwidth}
    \centering
\includegraphics[scale=0.18]{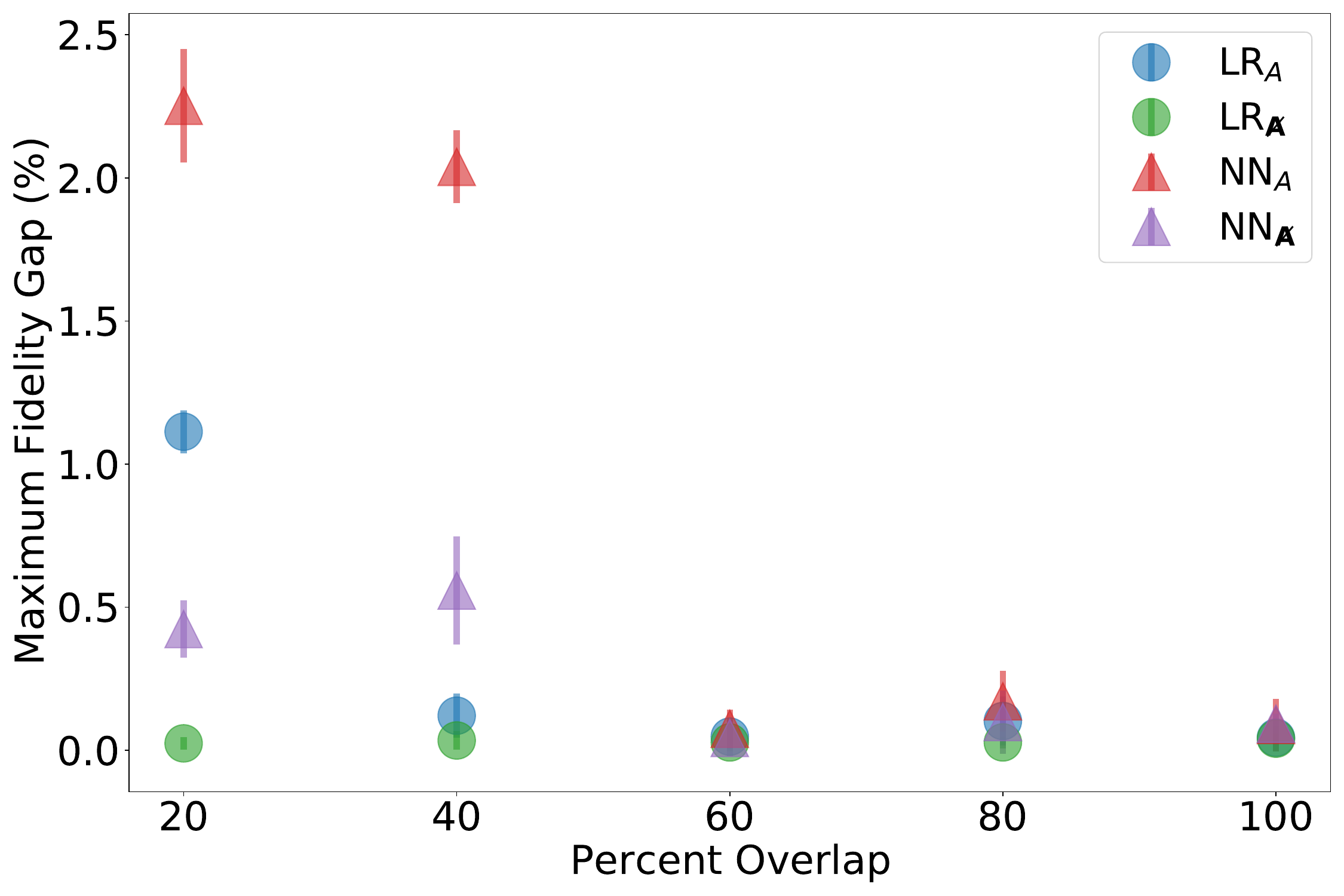}
    \caption{}
    \end{subfigure}
    \hfill
    \begin{subfigure}[b]{0.45\textwidth}
    \centering
    \includegraphics[scale=0.18]{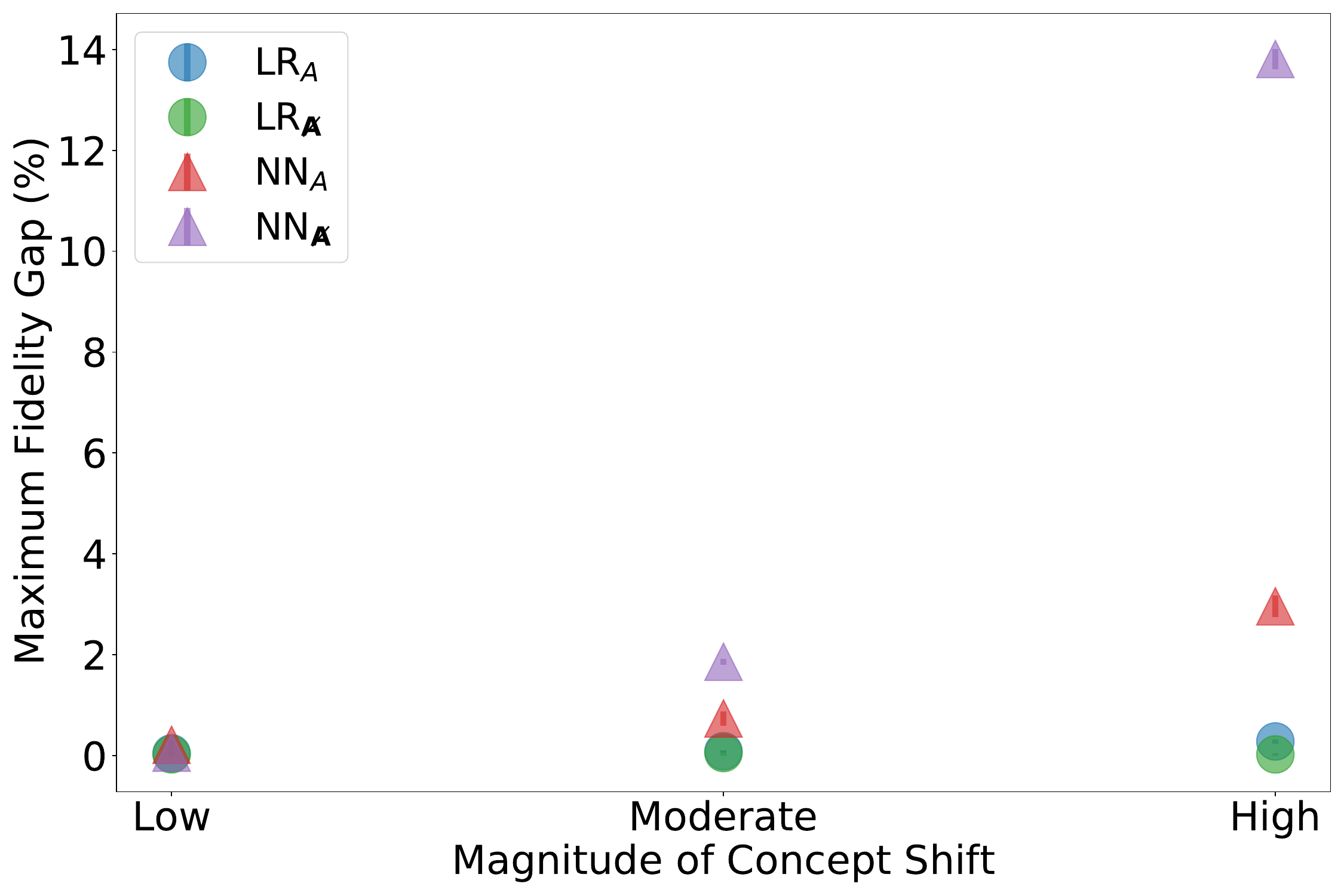}
    \caption{}
    \end{subfigure}
    \hfill
    \begin{subfigure}[b]{0.45\textwidth}
    \centering
    \includegraphics[scale=0.18]{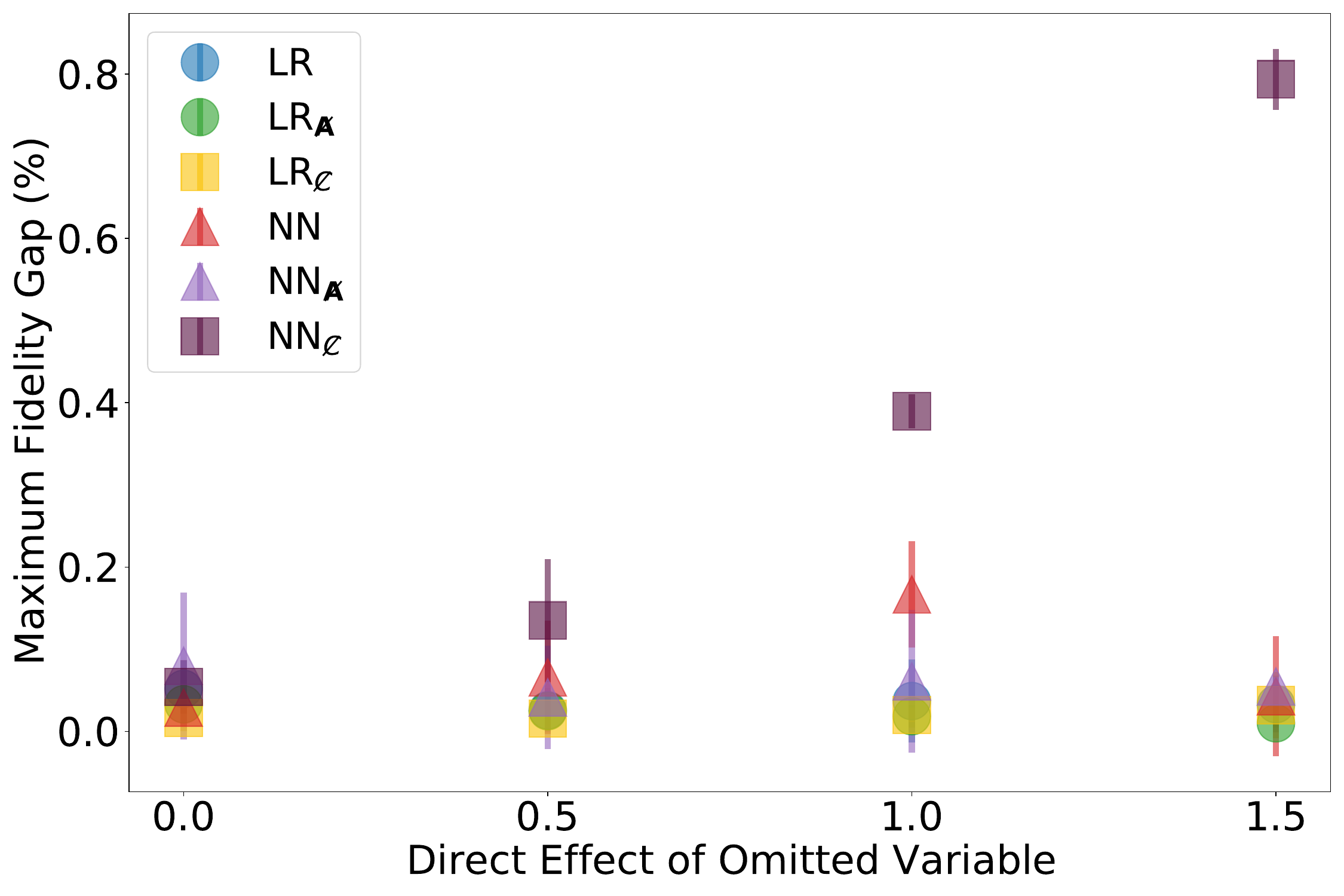}
    \caption{}
    \end{subfigure}
    \caption{Percent Maximum Fidelity Gap of LIME applied to models built on the synthetic datasets generated for (a) objective 1 - sample size, (b) objective 2 - covariate shift, (c) objective 3 - concept shift, and (d) objective 4 - omitted variables for LR with $A$, LR$_{A}$ in blue, LR without $A$, LR$_{\not \mathbf{A}}$ in green, NN with $A$, NN$_{A}$ in red, and NN without $A$, NN$_{\not \mathbf{A}}$ in violet, LR without $C$, LR$_{\not C}$ in yellow, and NN without $C$, NN$_{\not C}$ in plum. Circles represent linear models, and triangles represent neural network models.}
    \label{fig:appendix_results_sim_maximum}
\end{figure}

\begin{figure}[!htbp]
    \centering
    \begin{subfigure}[b]{0.45\textwidth}
        \centering
        \includegraphics[scale=0.2]{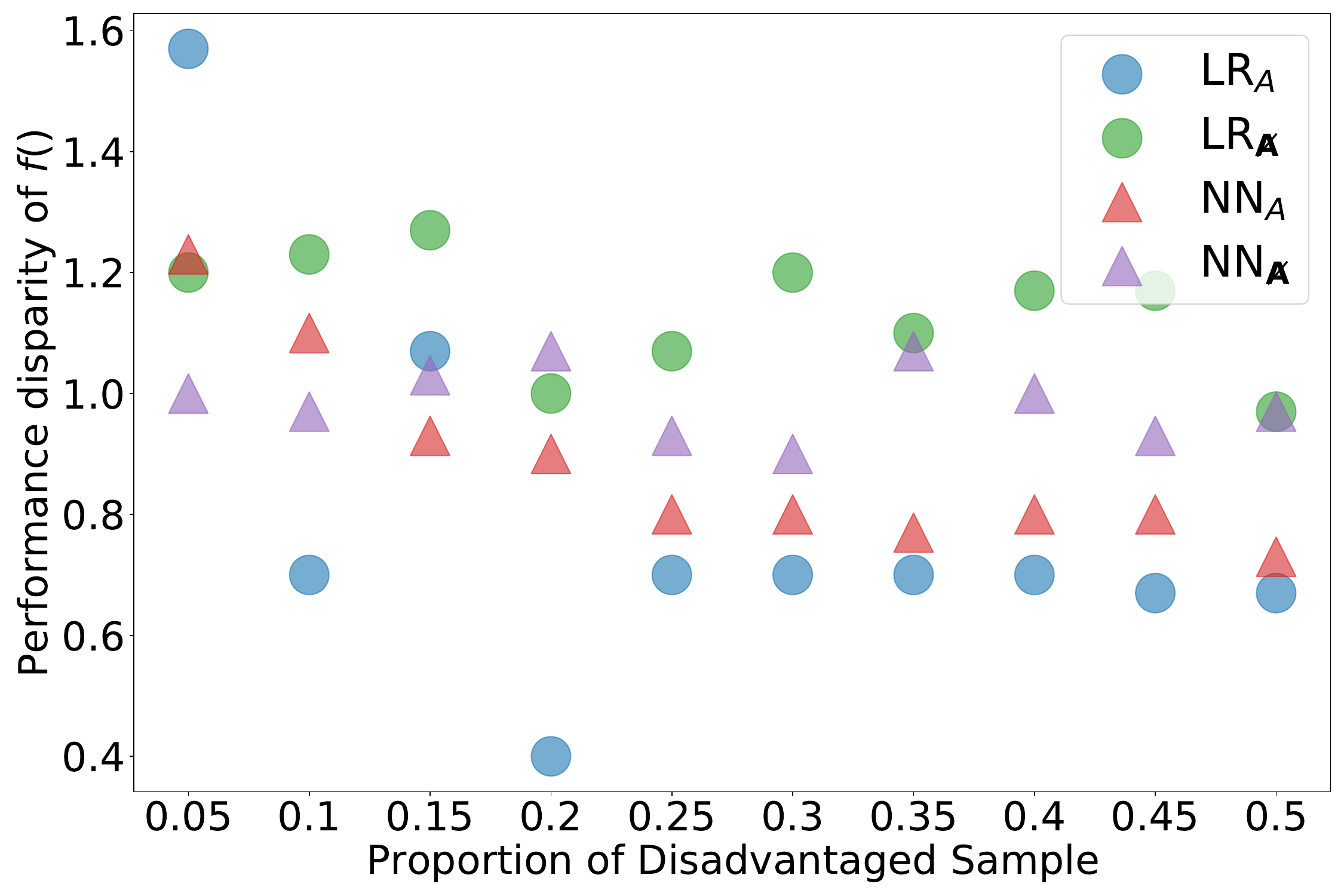}
        \caption{}
    \end{subfigure}
    \hfill
    \begin{subfigure}[b]{0.45\textwidth}
        \centering
        \includegraphics[scale=0.2]{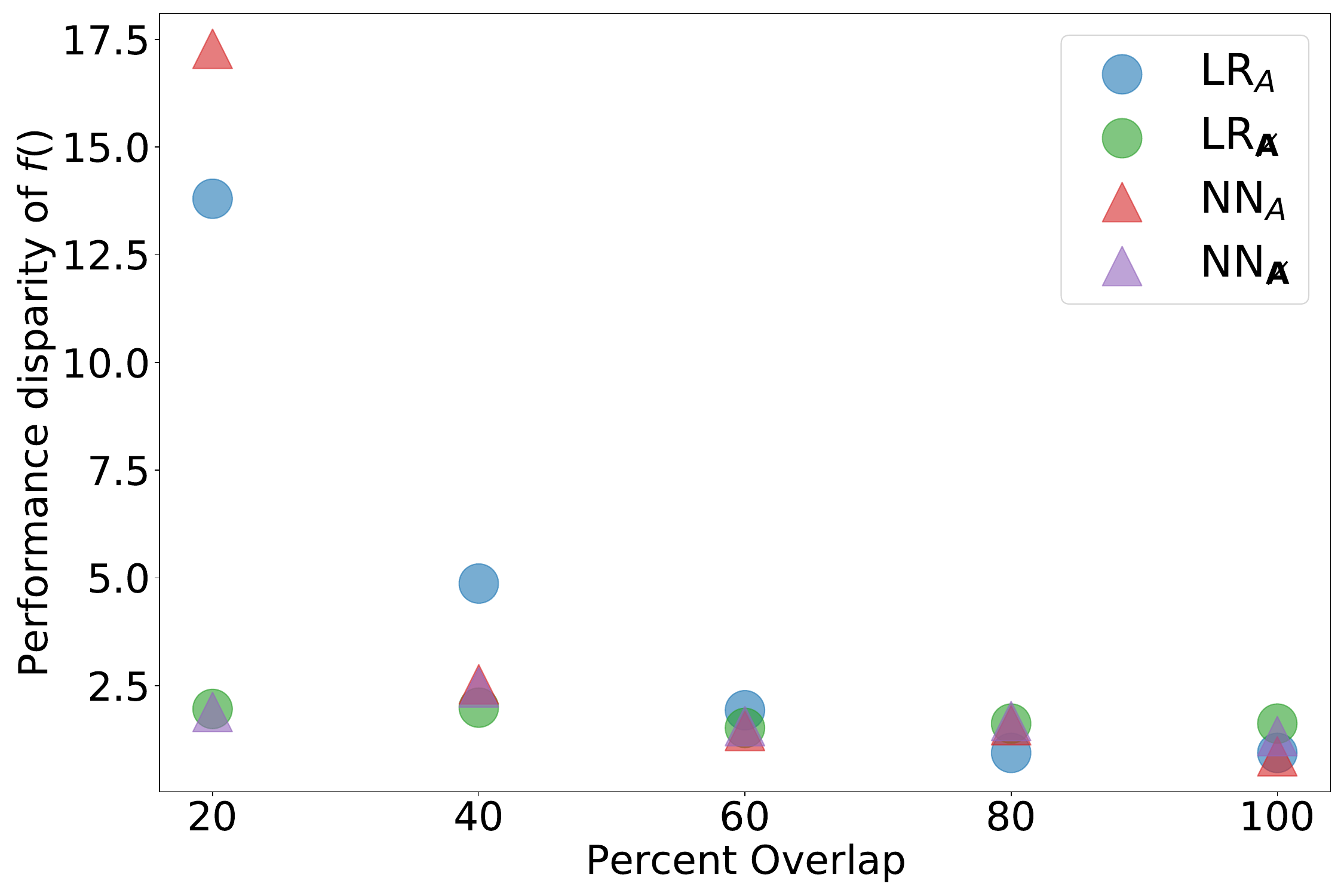}
        \caption{}
    \end{subfigure}
    \hfill
    \begin{subfigure}[b]{0.45\textwidth}
        \centering
        \includegraphics[scale=0.2]{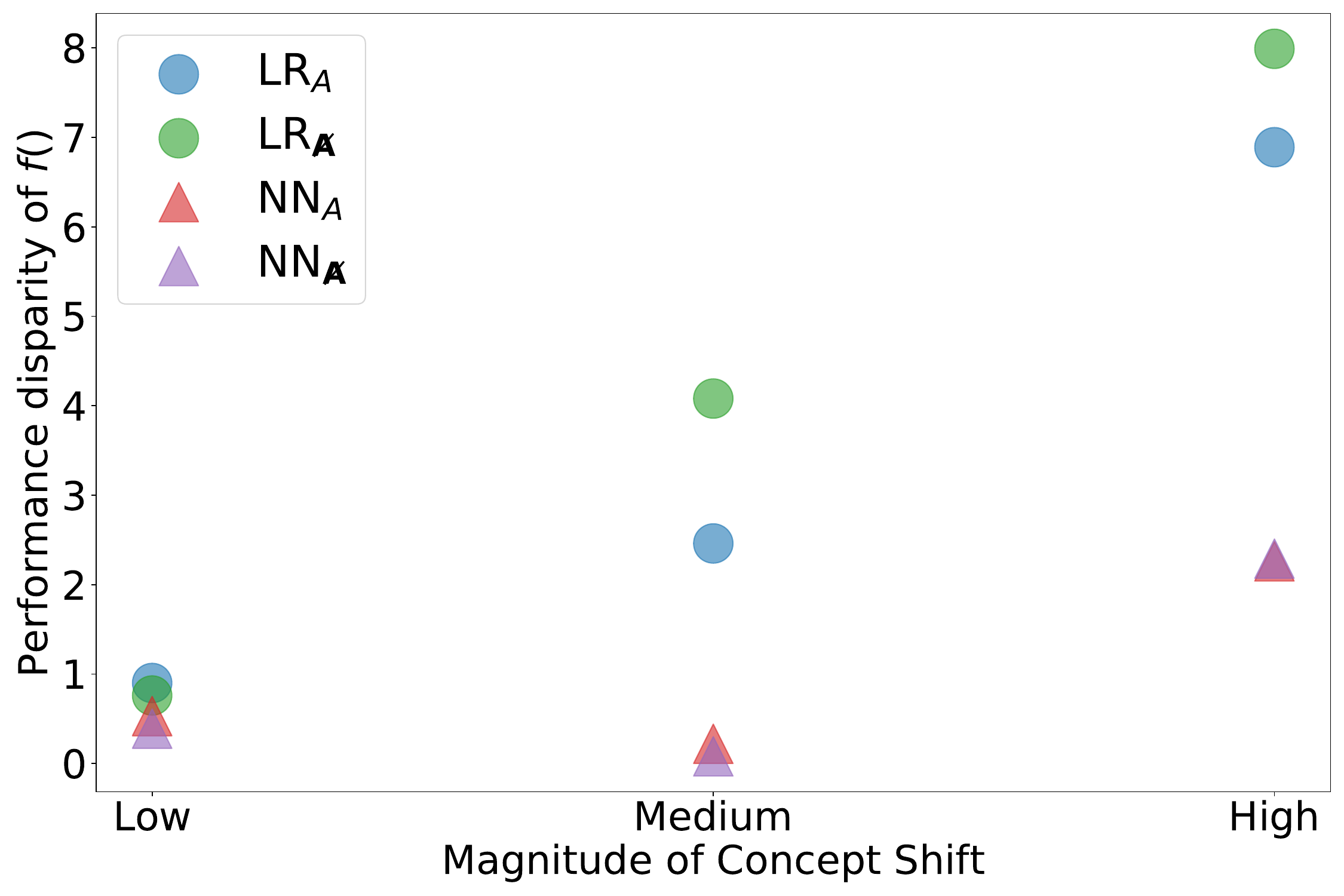}
        \caption{}
    \end{subfigure}
    \hfill
    \begin{subfigure}[b]{0.45\textwidth}
        \centering
        \includegraphics[scale=0.2]{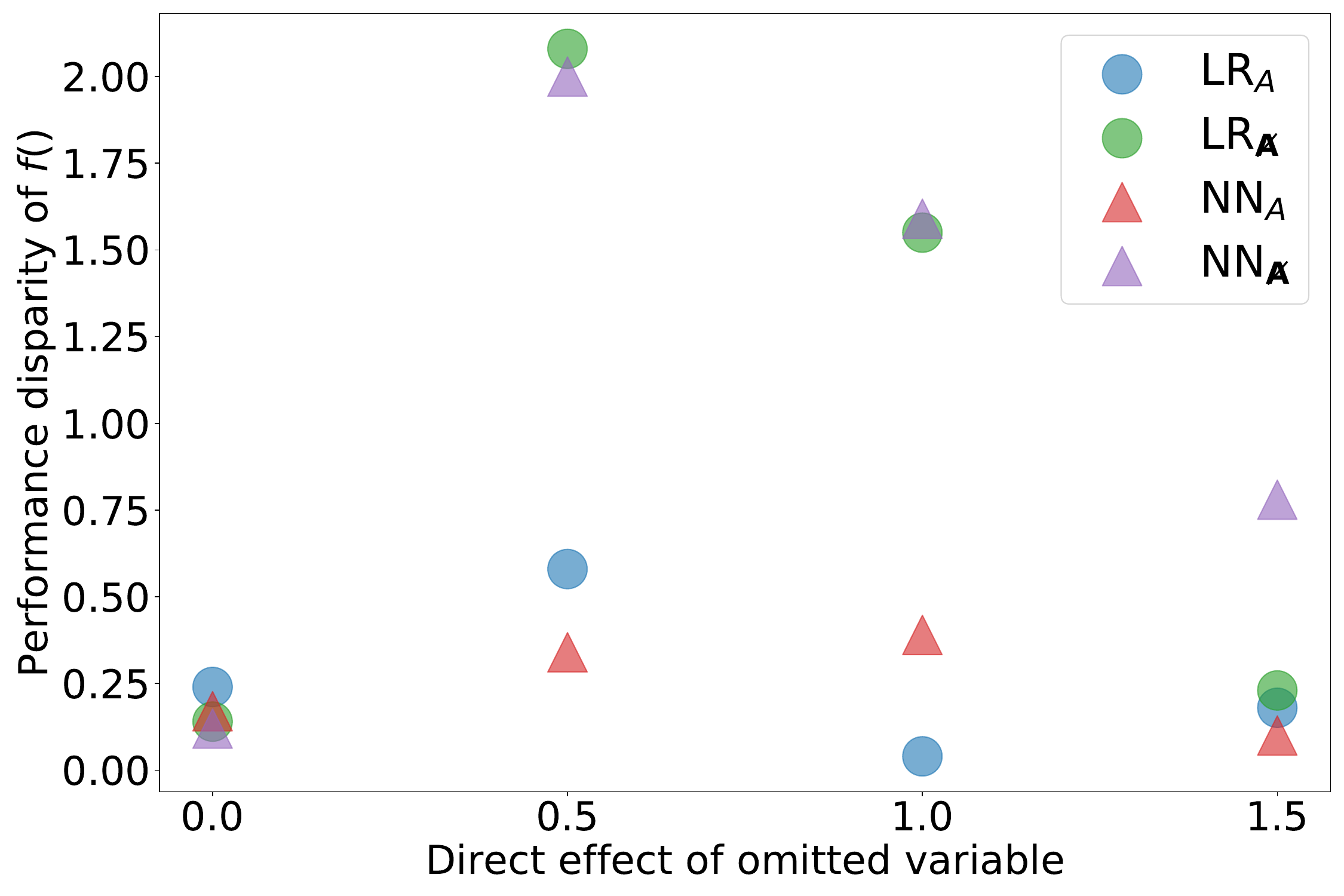}
        \caption{}
    \end{subfigure}
    \hfill
    \caption{Performance disparity of $f()$ calculated as Accuracy for A = 1 - Accuracy for A = 0 on the synthetic datasets generated with an increasing (a) proportion of the disadvantaged sample (objective 1), (b) overlap between the distribution of $L$ for $A=0$ between training and test distributions, (c) concept shift, and (d) direct effect of omitted variable $C$ for LR with $A$, LR$_{A}$ in blue, LR without $A$, LR$_{\not \mathbf{A}}$ in green, NN with $A$, NN$_{A}$ in red, and NN without $A$, NN$_{\not \mathbf{A}}$ in violet.  Circles represent linear models, and triangles represent neural network models.}
    \label{fig:bb_sim_1}
\end{figure}

\subsection{Detailed Results for Adult}
Here, we present results for Explanation Fidelity Metrics, $\Delta_{Acc}^{group}$ and $\Delta_{Acc}$ for objective 1 (varying percentage of the disadvantaged group, males in the training distribution) and objective 2 (varying overlap in the distribution of `hours-per-week,' $L$ for the disadvantaged group, males between the training and test distributions in Appendix Figures \ref{fig:appendix_results_adult_1} and \ref{fig:appendix_results_adult_2}, respectively.
Moreover,  we provide additional results for the Adult dataset for the black box model performance, a disparity in the prediction accuracy of the black box model between the advantaged and disadvantaged groups across all 4 objectives. 
We present these prediction disparities in Appendix Figures \ref{fig:bb_adut_1}(a), \ref{fig:bb_adut_1}(b) for objectives 1, 2 and in Tables 
 \ref{tab:adult_bb_3} and \ref{tab:adult_bb_4} for 3, and 4, respectively.

\begin{figure}[!htbp]
    \centering
    \begin{subfigure}[b]{0.45\textwidth}
    \centering
    \includegraphics[scale=0.18]{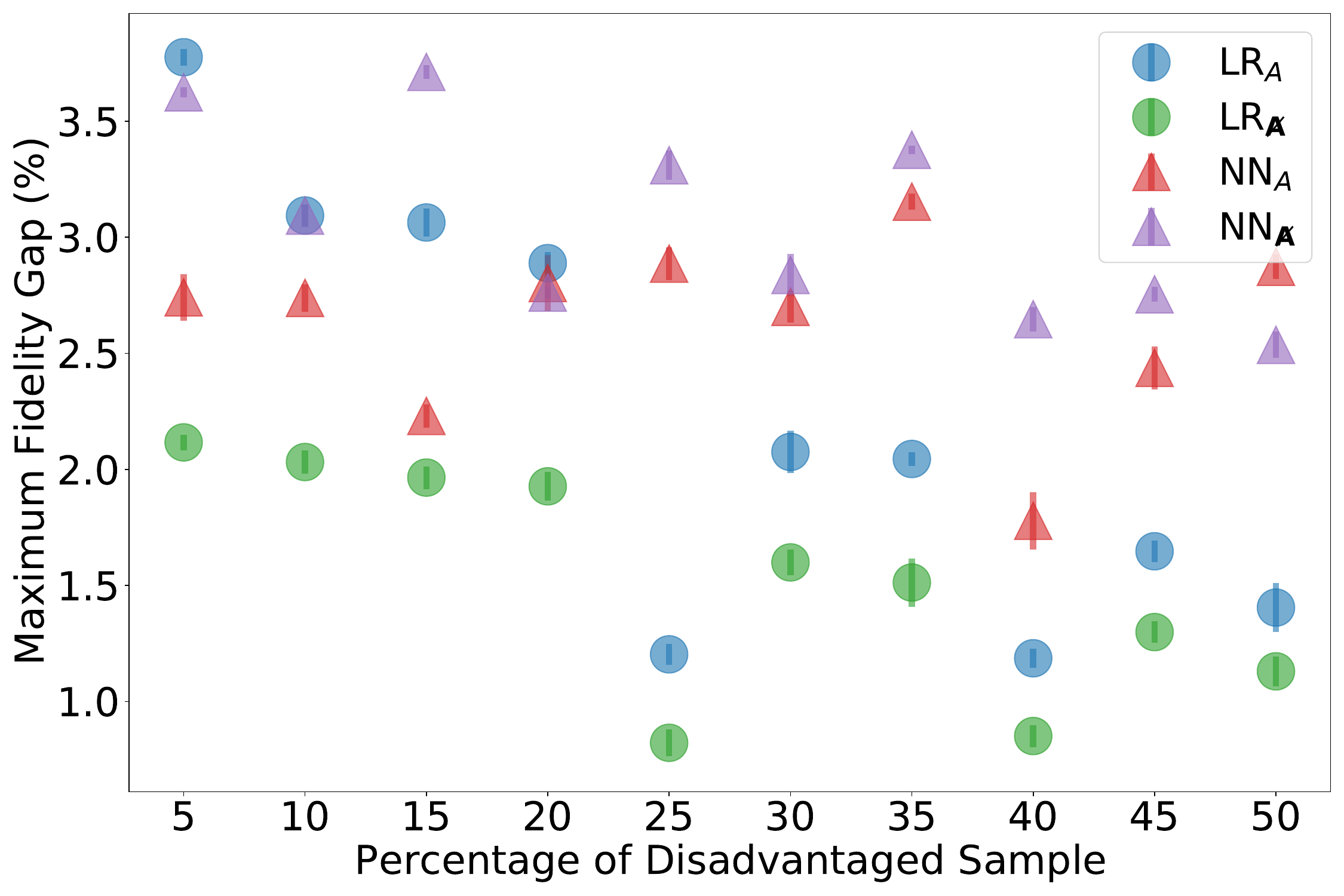}
    \caption{}
    \end{subfigure}
    \hfill
    \begin{subfigure}[b]{0.45\textwidth}
    \centering
\includegraphics[scale=0.18]{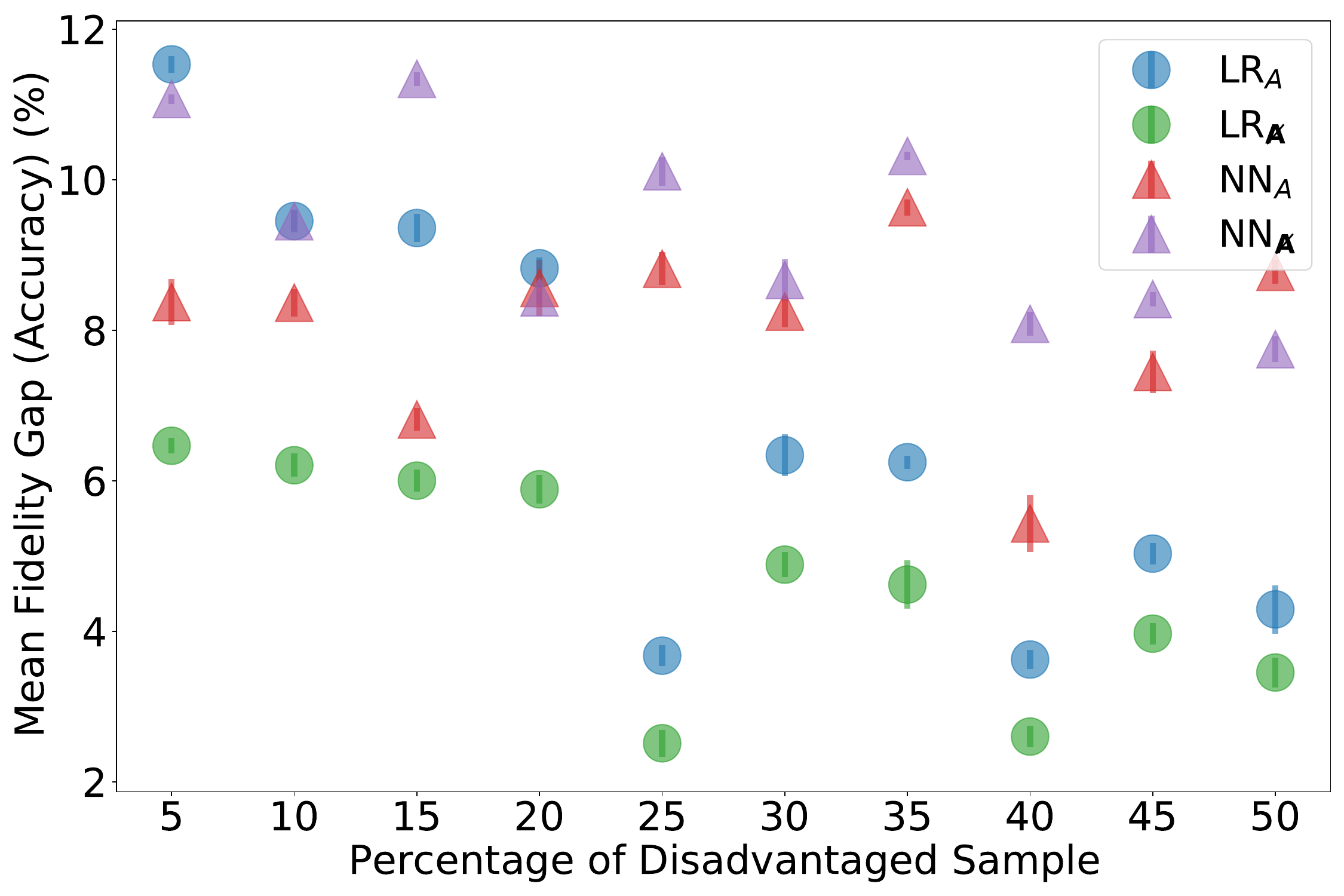}
    \caption{}
    \end{subfigure}
    \caption{(a) Percent Maximum Fidelity Gap, $\Delta_{Acc}$, (b) mean fidelity gap in accuracy, $\Delta_{Acc}^{group}$ 
    of LIME on Adult dataset with variation in the proportion of the `males' ($A$) in the training sample (objective 1) for LR with $A$, LR$_{A}$ in blue, LR without $A$, LR$_{\not \mathbf{A}}$ in green, NN with $A$, NN$_{A}$ in red, and NN without $A$, NN$_{\not \mathbf{A}}$ in violet.  Circles represent linear models, and triangles represent neural network models. }
    \label{fig:appendix_results_adult_1}
\end{figure}

\begin{figure}[h]
    \centering
    \begin{subfigure}[b]{0.45\textwidth}
    \centering
    \includegraphics[scale=0.18]{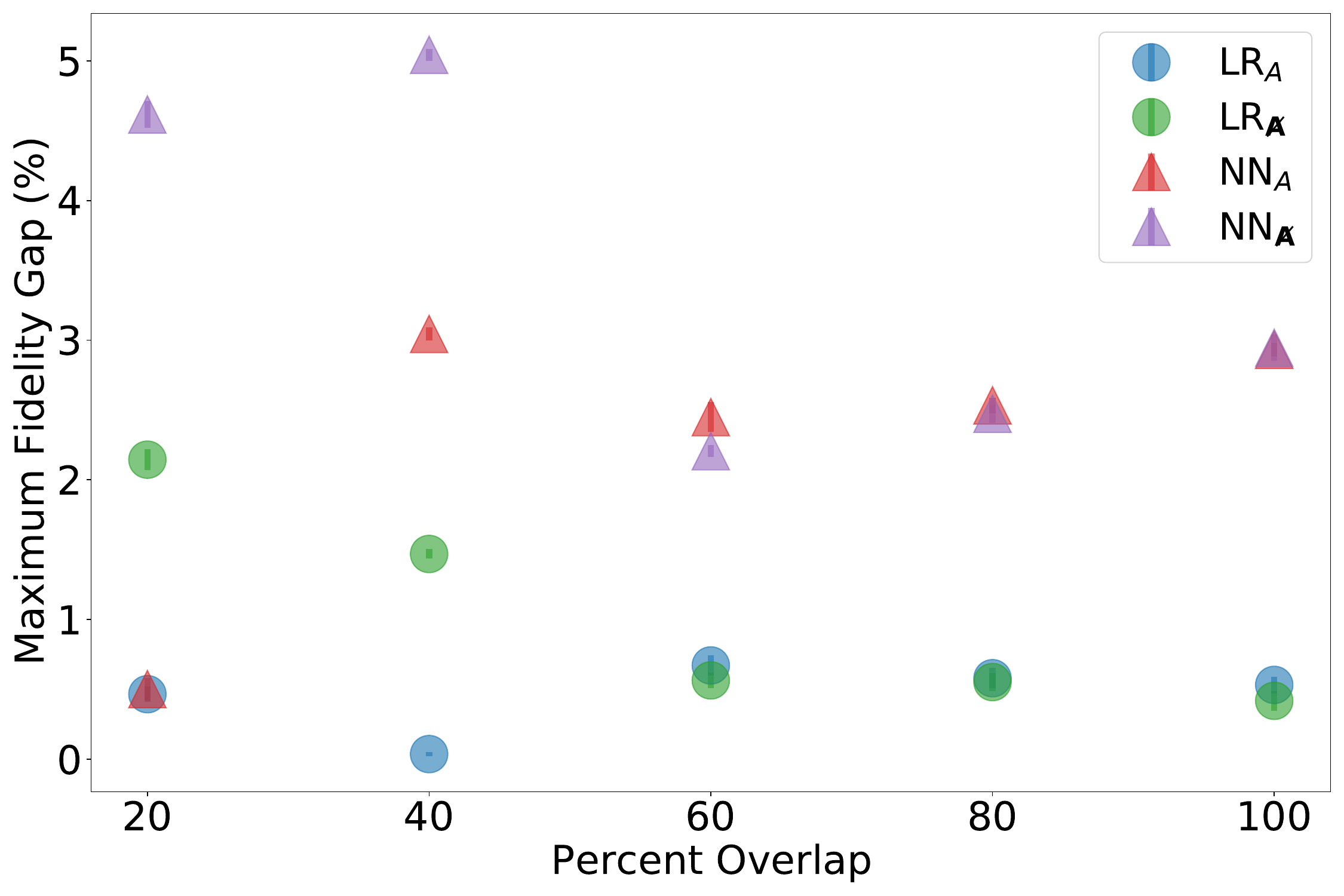}
    \caption{}
    \end{subfigure}
    \hfill
    \begin{subfigure}[b]{0.45\textwidth}
    \centering
\includegraphics[scale=0.18]{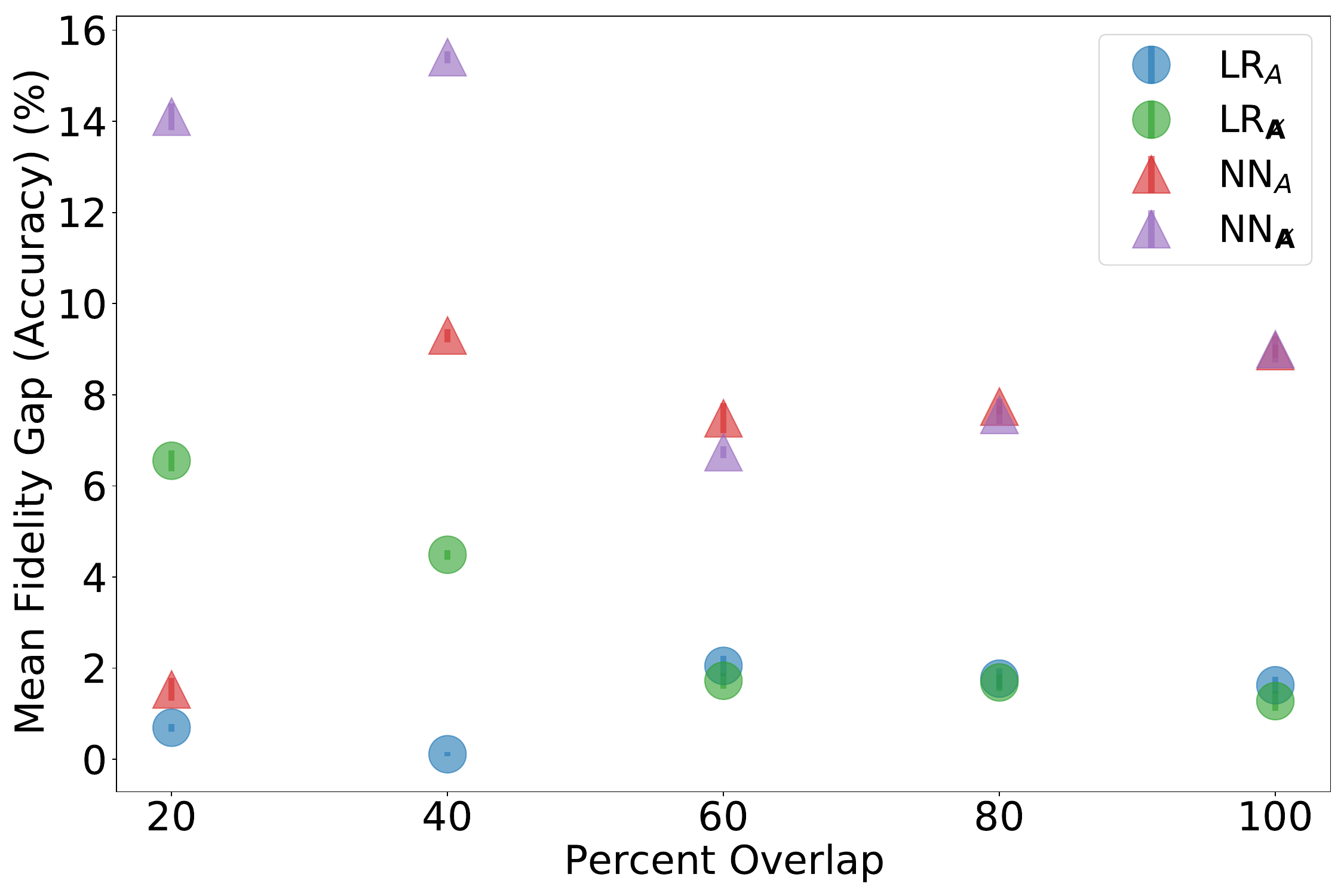}
    \caption{}
    \end{subfigure}
    \caption{(a) Percent Maximum Fidelity Gap, $\Delta_{Acc}$, (b) mean fidelity gap in accuracy, $\Delta_{Acc}^{group}$
    of LIME on the Adult dataset with variation in the overlap (covariate shift) in the distribution of the `males' ($A$) in the training sample and the test set (objective 2) for LR with $A$, LR$_{A}$ in blue, LR without $A$, LR$_{\not \mathbf{A}}$ in green, NN with $A$, NN$_{A}$ in red, and NN without $A$, NN$_{\not \mathbf{A}}$ in violet. Circles represent linear models, and triangles represent neural network models. }
    \label{fig:appendix_results_adult_2}
\end{figure}

\begin{figure}[!htbp]
    \centering
    \begin{subfigure}[b]{0.45\textwidth}
        \centering
        \includegraphics[scale=0.2]{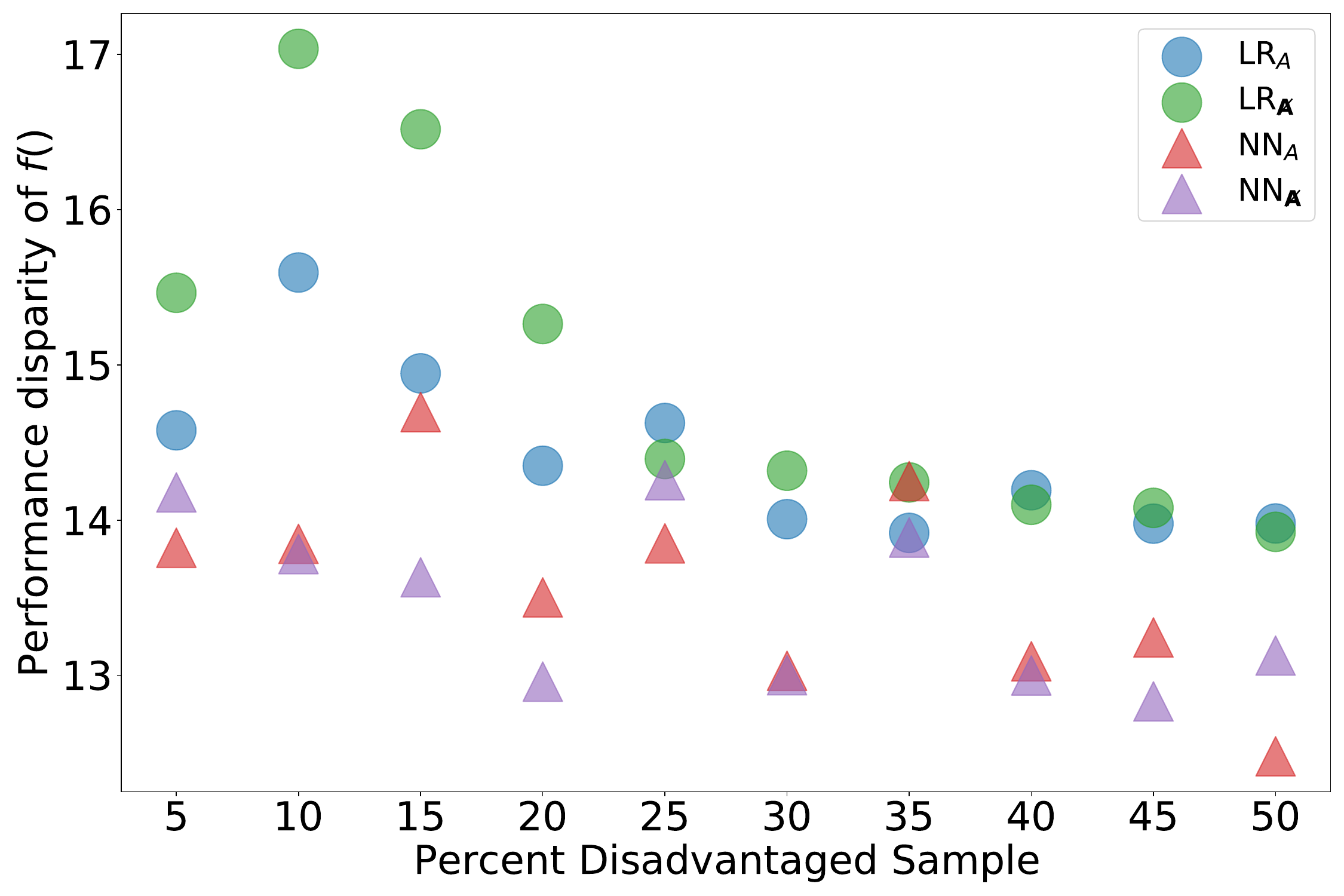}
        \caption{}
    \end{subfigure}
    \hfill
        \begin{subfigure}[b]{0.45\textwidth}
            \centering
            \includegraphics[scale=0.2]{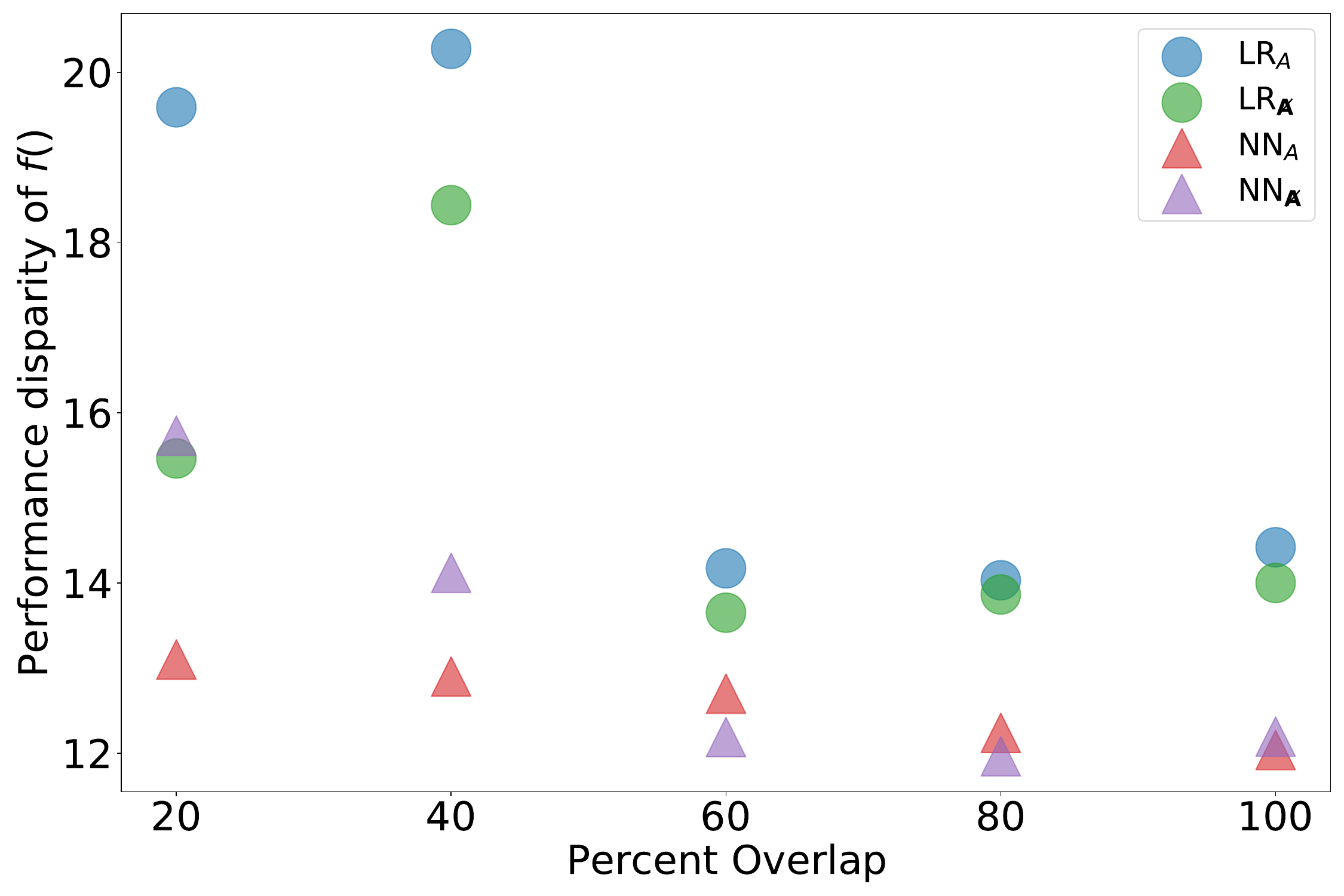}
            \caption{}
        \end{subfigure}
    \caption{Performance disparity of $f()$, black box model, calculated as Accuracy for A = 1 - Accuracy for A = 0 for Adult (a) with varying proportion of `male' group (objective 1), and (b) with varying overlap (covariate shift in `hours-per-week,' $L$) between train and test distribution for disadvantaged `male' group (objective 2). Circles represent linear models, and triangles represent neural network models.}
    \label{fig:bb_adut_1}
\end{figure}

\begin{table}[!htbp]
    \parbox{0.45\textwidth}{
        \centering
         \begin{tabular}{lccc}
        \toprule
        Model & Acc$_{A=0}$ & Acc$_{A=1}$ & $\mid \text{Acc}_{A=1} - \text{Acc}_{A=1}\mid$  \\ \midrule
        LR$_{A}$ & 76.41 & 90.21 & 13.80 \\ 
        LR$_{\not \mathbf{A}}$ & 76.30 & 90.50 & 14.21 \\ \hline
        NN$_{A}$ & 78.10 & 91.10 & 13.01 \\ 
        NN$_{\not \mathbf{A}}$ & 78.30 & 91.50 & 13.22 \\ \bottomrule
    \end{tabular}
    \caption{Black box model performance with respect to percentage accuracy for disadvantaged ($A=0$) and advantaged ($A=1$) groups with the difference in accuracy across groups for LR$_{A}$, LR$_{\not \mathbf{A}}$, NN$_{A}$, NN$_{\not \mathbf{A}}$ for concept shift between `hours-per-week' and `income' for male group for Adult (objective 3)
    \label{tab:adult_bb_3}}
    }
    \hfill
    \parbox{0.45\textwidth}{
        \centering
        \begin{tabular}{lccc}
    \toprule
     Model & Acc$_{A=0}$ & Acc$_{A=1}$ & $\mid \text{Acc}_{A=1} - \text{Acc}_{A=1}\mid$  \\ \midrule
        LR$_{C}$ &  76.36 & 90.69 & 14.33 \\ 
        LR$_{\not C}$  & 76.64 & 90.87& 14.23\\ \hline
        NN$_{C}$  & 78.78 & 91.33 & 12.55\\ 
        NN$_{\not C}$ & 79.00 & 91.92 & 12.92 \\
        \bottomrule
    \end{tabular}
    \caption{Black box model performance with respect to percentage accuracy for disadvantaged ($A=0$) and advantaged ($A=1$) groups with the difference in accuracy across groups for LR with `Nationality' included LR$_C$, LR with `Nationality' excluded LR$_{\not C}$, NN with `Nationality' included NN$_C$, and NN with `Nationality' excluded NN$_{\not C}$ for Adult (objective 4).}
    \label{tab:adult_bb_4}}
\end{table}

\end{document}